\documentclass[10pt,twocolumn,letterpaper]{article}

\usepackage{cvpr}              
\usepackage{gensymb}
\usepackage{multirow}

\definecolor{cvprblue}{rgb}{0.21,0.49,0.74}
\definecolor{GammaColor}{rgb}{0.5,0,0.7}

\usepackage[pagebackref,breaklinks,colorlinks,allcolors=cvprblue]{hyperref}

\def\paperID{17024} 
\def\confName{CVPR}
\def\confYear{2025}

\title{RDD: Robust Feature Detector and Descriptor using Deformable Transformer}

\author{
    Gonglin Chen$^{1, 2}$,
    Tianwen Fu$^{1, 2}$,
    Haiwei Chen$^{1, 2}$, 
    Wenbin Teng$^{1, 2}$,
    Hanyuan Xiao$^{1, 2}$, 
    Yajie Zhao$^{1, 2}$\thanks{Corresponding author.}
    \and
    $^{1}$Institute for Creative Technologies
    \and
    $^{2}$University of Southern California
    \and
    {\tt\small \{gonglinc, tianwenf\}@usc.edu, chenh@ict.usc.edu},\\
    {\tt\small \{wenbinte, hanyuanx\}@usc.edu},
    {\tt\small zhao@ict.usc.edu}
}

\newcommand{\shortname}{RDD }
\begin{document}
\maketitle
\begin{abstract}
As a core step in structure-from-motion and SLAM, robust feature detection and description under challenging scenarios such as significant viewpoint changes remain unresolved despite their ubiquity. While recent works have identified the importance of local features in modeling geometric transformations, these methods fail to learn the visual cues present in long-range relationships. We present \textbf{Robust Deformable Detector} (RDD), a novel and robust keypoint detector/descriptor leveraging the deformable transformer, which captures global context and geometric invariance through deformable self-attention mechanisms. Specifically, we observed that deformable attention focuses on key locations, effectively reducing the search space complexity and modeling the geometric invariance. Furthermore, we collected an Air-to-Ground dataset for training in addition to the standard MegaDepth dataset. Our proposed method outperforms all state-of-the-art keypoint detection/description methods in sparse matching tasks and is also capable of semi-dense matching. To ensure comprehensive evaluation, we introduce two challenging benchmarks: one emphasizing large viewpoint and scale variations, and the other being an Air-to-Ground benchmark — an evaluation setting that has recently gaining popularity for 3D reconstruction across different altitudes. Project page: \href{https://xtcpete.github.io/rdd/}{https://xtcpete.github.io/rdd/}.
\vspace{-0.5 em}

\begin{figure}
    \centering
    \includegraphics[width=0.99\linewidth]{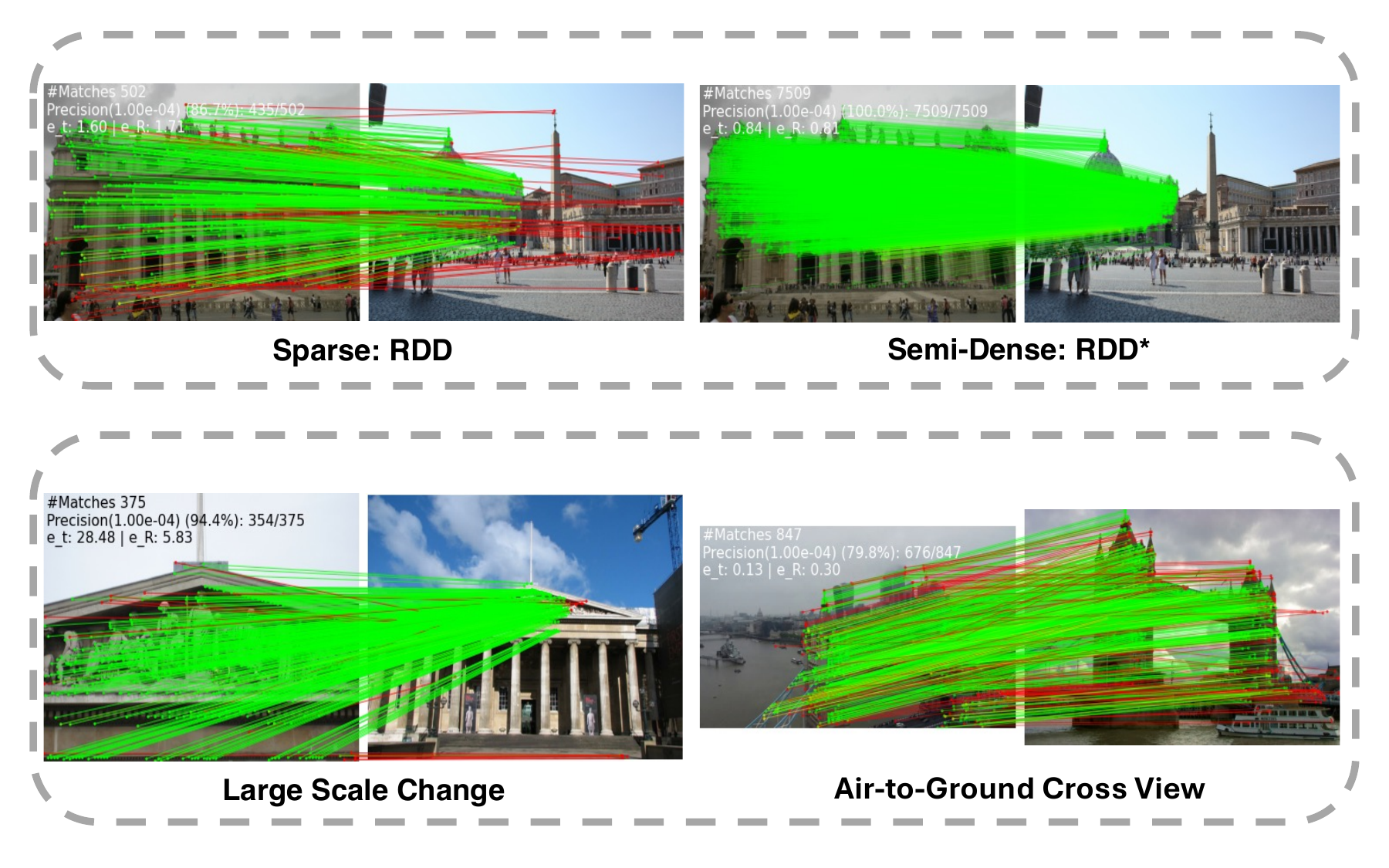}
    \vspace{-1 em}
    \caption{Our proposed method effectively performs both sparse and semi-dense feature matching, referred to as \shortname and RDD*, respectively, as shown in the top section. \shortname demonstrates its ability to extract accurate keypoints and robust descriptors, enabling reliable matching even under significant scale and viewpoint variations, as illustrated in the bottom section.}
    \label{fig:teaser}
    \vspace{-2 em}
\end{figure}

\end{abstract}
\vspace{-1 em}
\section{Introduction}



Keypoint detection and description are central to numerous 3D computer vision tasks, including structure-from-motion (SfM), visual localization, and simultaneous localization and mapping (SLAM). These tasks depend on reliable keypoints and descriptors to establish accurate tracks, which are crucial for downstream algorithms to interpret spatial relationships—a requirement in fields from robotics to augmented reality~\cite{kerbl3Dgaussians}. When applied to real-world tasks, these systems face challenges such as significant viewpoint shifts, lighting variations, and scale changes.

Historically, feature detection and description methods relied on rule-based techniques such as SIFT , SURF, and ORB~\cite{Lowe:2004:DIF:993451.996342,SURF2008, ORB2011}. However, these methods oftenf struggled with the aforementioned challenges. To better address these issues, recent works~\cite{superpoint2018, gleize2023silksimplelearned, tyszkiewicz2020disk, Zhao2022ALIKE} leverage deep neural networks based on data-driven approaches which often extract more robust and discriminative descriptors than the rule-based ones. Despite the enhanced robustness that learning-based methods have, they often rely on regular convolutions to encode images~\cite{superpoint2018, tyszkiewicz2020disk, gleize2023silksimplelearned, potje2024cvpr, Zhao2022ALIKE}, which ignore the geometric invariance and long-distance awareness critical for robust description under challenging conditions. Subsequently,~\cite{luo2020aslfeat, LIFT2016, liu2019giftlearningtransformationinvariantdense} have partially addressed geometric invariance by estimating the scale and orientation of keypoint descriptors to model affine transformations. ALIKED~\cite{zhao2023aliked} and ASLFeat~\cite{luo2020aslfeat} advanced this approach by using deformable convolutions capable of modeling any geometric transformations. However, these methods are limited to modeling transformations within local windows; features learned with the localized kernels of convolution operators fall short of learning important visual cues that depend on long-range relationships (e.g. vanishing lines).



In this paper, we focus on feature detection and description in challenging scenarios that haven't been effectively addressed in prior works -- specifically, extracting reliable and discriminative keypoints and descriptors across images with large camera baselines, significant illumination variations, and scale changes. We therefore propose RDD: a novel two-branch architecture for keypoint detection and description that can model both geometric invariance and global context at the same time. In particular, we design two dedicated network branches: a fully convolutional architecture to perform keypoint detection and a transformer-based architecture for extracting descriptors: the objectives of keypoint detection and description have previously been found not to align perfectly~\cite{li2022decoupling}. We observe that the convolutional neural network excels in detecting sub-pixel keypoints thanks to its expressiveness, whereas a transformer-based architecture is particularly suited for learning features that captures global context and possess geometric invariance through its self-attention mechanism~\cite{vaswani2017attention, dosovitskiy2021imageworth16x16words}. However, as self-attention layers attend to all spatial locations in the image, they significantly increase computational overhead and may reduce descriptor discriminability. To address this, we have adopted the deformable attention~\cite{zhu2020deformable} for keypoint description, allowing our designed network to selectively focus on key locations, which greatly reduces time complexity while maintaining the capability to learn geometric invariance and global context.

The two-branch architecture of \shortname allows specialized architectures to learn keypoint detection and description independently. Our experiments have found that such design choice accelerates convergence and leads to better overall performance, as proven in~\cref{sec:ablation}. Overall, \shortname is designed to be robust, achieving competitive performance on challenging imagery and also capable of semi-dense matching~\cref{fig:teaser}. This enhanced capability improves accuracy and stability in 3D vision applications, such as structure-from-motion (SfM)~\cite{chen2024geometryawarefeaturematchinglargescale} and relative camera pose estimation under difficult imaging conditions.
Our approach outperforms current state-of-the-art methods for keypoint detection and description on standard benchmarks, including MegaDepth-1500~\cite{dusmanu2019d2net}, HPatches~\cite{hpatches_2017_cvpr}, and Aachen-Day-Night~\cite{sattler2018benchmarking6dofoutdoorvisual}, across various tasks. However, to the best of our knowledge, no existing benchmarks accurately capture the aftermentioned challenging scenarios. To further validate the robustness of our method and provide a more comprehensive evaluation, we have collected and derived two additional benchmark datasets. The main contributions of our work are:

\begin{itemize} 
    \item A novel two branch architecture that utilizes a convolutional neural network for detecting accurate keypoints and transformer based network to extract robust descriptors under challenging scenarios, all while maintaining efficient runtime performance. 
    \item An innovative refinement module for semi-dense matching, improving matching accuracy and density.
    \item Dataset contribution: MegaDepth-View, derived from MegaDepth~\cite{li2018megadepth}, specifically focuses on challenging image pairs with large viewpoint and scale variations. Air-to-Ground dataset provides diverse aerial and ground cross view imagery for training and evaluation, enhancing robustness and performance and providing a new evaluation setting.
\end{itemize}

\begin{figure*}
  \centering
  \vspace{-2em}
   \includegraphics[width=0.94\linewidth]
   {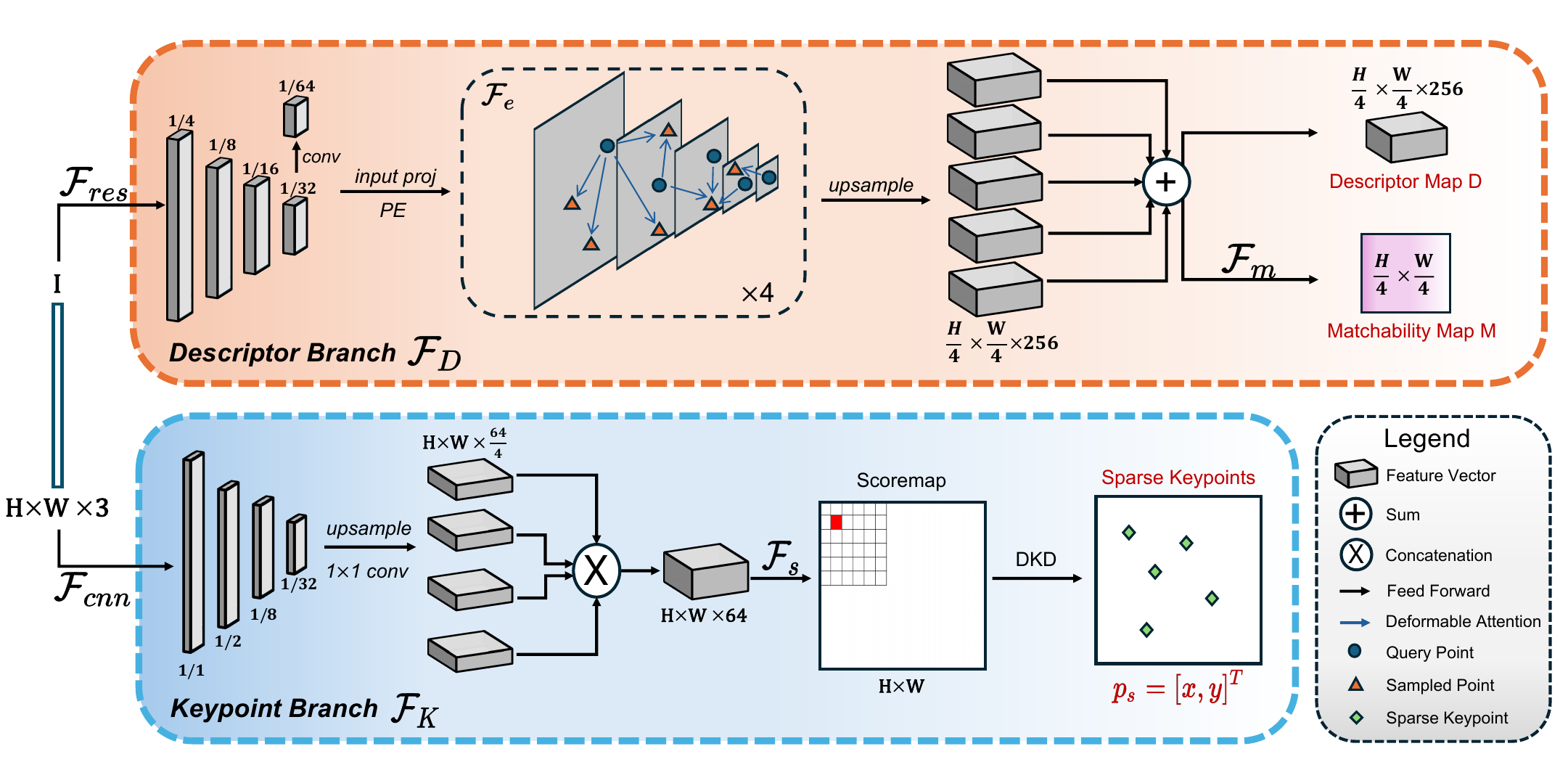}
   \vspace{-0.5 em}
   \caption{An overview of our network architecture. Descriptor Branch $\mathcal{F}_D$ and Keypoint Branch $\mathcal{F}_K$ process an input image $I \in \mathbb{R}^{H \times W \times 3}$ independently.  \textbf{Descriptor Branch:} 4 layers of multiscale feature maps $\{\mathbf{x}_{res}^l\}_{l=1}^{L}$ are extracted by passing $I$ through ResNet-50~\cite{he2015deepresiduallearningimage} $\mathcal{F}_{res}$. Additional feature map at scale of $1/64$ is added by applying a simple CNN on the last feature map and then they are feed to a transformer encoder $\mathcal{F}_e$ with positional embeddings~\cite{vaswani2017attention}. We up-sample all feature maps output by $\mathcal{F}_e$ to size $H/k \times W/k$ where $k=4$ is the patch size. Feature maps are then summed together to generate the dense descriptor map $D$. A classification head $\mathcal{F}_m$ is applied to the descriptor map to estimate a matchability map $M$.
   \textbf{Keypoint Branch:} $I$ passes through a lightweight CNN with residual connection~\cite{he2015deepresiduallearningimage} $\mathcal{F}_{cnn}$ to capture multi-scale features $\{\mathbf{x}_{cnn}^l\}_{l=1}^{L}$. Features are then up-sampled to size $H \times W$ and concatenated to generate a feature map of $H \times W \times 64$. A score map $S$ is estimated by a classification head $\mathcal{F}_s$. Final sub-pixel keypoints are detected using DKD~\cite{Zhao2022ALIKE}.}
   \label{fig:pipeline}\vspace{-1 em}
\end{figure*}

\vspace{-0.2em}
\section{Related Work}
\subsection{Geometric Invariant Descriptors}
In geometric invariance modeling, previous methods have primarily focused on two key aspects: scale and orientation. In traditional hand-crafted methods, SIFT~\cite{siftflow} estimates a keypoint's scale and orientation by analyzing histograms of image gradients and subsequently extracts and normalizes image patches around the keypoint to construct scale- and orientation-invariant descriptors. Similarly, ORB~\cite{ORB2011} computes orientation based on the keypoint’s center of mass, followed by rotation of the image patches to ensure orientation invariance.

Learning-based methods approach geometric invariance through two primary strategies. The majority of these methods~\cite{superpoint2018, tyszkiewicz2020disk, dusmanu2019d2net, r2d2, gleize2023silksimplelearned, potje2024cvpr} relies heavily on data augmentation techniques to achieve scale and orientation invariance. On the other hand, some learning-based approaches explicitly model scale and orientation. For instance, LIFT~\cite{LIFT2016} mimics SIFT~\cite{siftflow} by detecting keypoints, estimating their orientation, and extracting descriptors with different neural networks. Other methods, such as AffNet~\cite{mishkin2018repeatabilityenoughlearningaffine} and LF-Net~\cite{ono2018lfnetlearninglocalfeatures}, predict affine transformation parameters and apply these transformations via Spatial Transformer Networks~\cite{jaderberg2016spatialtransformernetworks} to produce affine-invariant descriptors. These methods generally assume a predefined geometric transformation, typically affine transformations. More recent advancements, such as ASLFeat~\cite{luo2020aslfeat} and ALIKED~\cite{zhao2023aliked}, employ deformable convolutions to directly extract geometric-invariant features, offering a more flexible and adaptive approach to handling geometric variations.

Inspired by the use of deformable convolutions in feature description, the proposed network \shortname employs deformable attention to extract geometric invariant features that are not restricted by local windows.

\subsection{Decoupling Keypoint and Descriptor Learning}

In traditional handcrafted methods for feature detection and description, such as SIFT~\cite{siftflow} and ORB~\cite{ORB2011}, the detection and description processes are performed independently. This decoupled approach was also common in early learning-based methods~\cite{barrosolaguna2019keynetkeypointdetectionhandcrafted, mishchuk2018workinghardknowneighbors}, where keypoint detection and description were treated as separate tasks. However, later works~\cite{dusmanu2019d2net, superpoint2018, gleize2023silksimplelearned, r2d2} proposed an end-to-end framework, jointly optimizing keypoint detection and descriptor extraction. This joint learning approach ensures that the detected keypoints and their descriptors complement each other.
Despite the advantages, ~\cite{li2022decoupling} highlighted a potential drawback of joint training, observing that poorly learned descriptors could negatively impact keypoint detection quality. Subsequently, DeDoDe\cite{edstedt2024dedode} demonstrated that decoupling keypoint detection and description into two independent networks can significantly improve performance. Similarly, XFeat~\cite{potje2024cvpr} showed that using task-specific networks provides a better balance between efficiency and accuracy.
Inspired by these recent insights, \shortname consists of two separate branches for independent keypoint detection and description. We also formulate a training strategy that optimize the descriptor branch first and then the keypoint branch. This design enables both sparse and dense matching and ensures optimized performance across various scenarios.

\subsection{Efficient Attention Mechanisms}
The quadratic complexity of multi-head attention~\cite{vaswani2017attention} leads to long training schedules and limited spatial resolution for tractability. For 2D images, LR-Net~\cite{hu2019local} proposed a local relation layer, computing attention in a fixed local window. Linear Transformer~\cite{katharopoulos2020transformersrnnsfastautoregressive} proposes to reduce the computation complexity of vanilla attention to linear complexity by substituting the exponential kernel with an alternative kernel function. Deformable attention~\cite{zhu2020deformable} further allows the attention window to depend on the query features. With a linear projection predicting a deformable set of displacements from the center and their weight, each pixel in the feature map may attend to a fixed number of pixels of arbitrary distance. Deformable attention~\cite{zhu2020deformable} is well-suited for our objectives, enabling efficient learning of global context, making it a fitting choice for our approach.

\vspace{-0.5 em}
\section{Methods}
\label{methods}

Our method extracts descriptors and keypoints separately using a descriptor branch and a keypoint branch, as shown in~\cref{fig:pipeline}. For an input image $I\in\mathbb{R}^{H\times W \times 3}$, the descriptor branch first extracts multi-scale feature maps $\{x_{res}^l\}_{l=1}^{L}$ from $I$ using ResNet-50~\cite{he2015deepresiduallearningimage}, denoted as $\mathcal{F}_{res}$, and then estimates a descriptor map $D\in \mathbb{R}^{H/k \times W/k}$ through a deformable transformer encoder~\cite{zhu2020deformable}, denoted as $\mathcal{F}_{e}$, where $k=4$ is the patch size. The matchability map $M \in \mathbb{R}^{H/k \times W/k}$ is then estimated from $D$ with a classification head $\mathcal{F}_{m}$. The keypoint branch $\mathcal{F}_K$ estimates a scoremap $S \in \mathbb{R}^{H\times W}$ by passing image $I$ through a lightweight CNN with residual connection~\cite{he2015deepresiduallearningimage} $\mathcal{F}_{cnn}$ and a classification head $\mathcal{F}_s$, then sub-pixel keypoints $p_s = [x, y]^T$ are detected with differentiable keypoint detection (DKD)~\cite{Zhao2022ALIKE}. 

The detected keypoints are used to sample their corresponding descriptors $d \in \mathbb{R}^{256}$ from $D$. In~\cref{sec:prelim}, we provide an overview of DKD~\cite{Zhao2022ALIKE} and deformable attention~\cite{zhu2020deformable}. In~\cref{sec:architecture}, we introduce the network architecture. In~\cref{sec:matching}, we discuss how sparse and dense matching are obtained. Finally, the training details and loss functions are presented in~\cref{sec:implementation}.

\subsection{Preliminaries}
\label{sec:prelim}

\subsubsection{Differentiable Keypoint Detection} 

Given the scoremap $S$, DKD detects sub-pixel keypoint locations through partially differentiable operations. First, local maximum scoremap $S_{nms}$ is obtained by non-maximum suppression in local $N \times N$ windows. Pixel-level keypoints $ p_{nms} = [x', y']^T $ are then extracted by applying a threshold to $S_{nms}$. DKD then selects all scores $s(i, j)$ in the $N \times N$ windows centered on $p_{nms}$ from $S$ and estimates a sub-pixel-level offset through following steps:

$s(i, j)$ are first normalized with softmax:
\begin{equation}	
    s'(i,j) = \mathbf{softmax} \left( \frac{s(i,j)-s_{max}}{t_{det}} \right),
\label{eq:softmax}	
\end{equation}
where $t_{det}$ is the temperature. Then $s'(i,j)$ represents the probability of $[i,j]^T$ to be a keypoint within the local window. Thus, the expected position of the keypoint in the local window can be given by integral regression~\cite{sun2018integralhumanposeregression, gu2021removing}:
\begin{equation}
    \label{equ_softlocal}
    [\hat i, \hat j]_{soft}^T = \sum_{0\le i,j<N} s'(i,j)[i,j]^T.
\end{equation}	
The final estimated sub-pixel-level keypoint is
\begin{equation}
    \boldsymbol{p_s} = [x, y]^T = [x',y']^T + [\hat i, \hat j]_{soft}^T .
     \label{eq:dkd}
\end{equation}


\subsubsection{Deformable Attention}

Given a feature map $\mathbf{x} \in \{\mathbf{x}_{res}^l\}_{l=1}^{L}$, deformable attention selects a small set of sample keys $\mathbf{p}_q$ corresponding to the query index $q$ and its feature vectors $\mathbf{z}_q$. The deformable attention feature $\mathbf{x}_d$ is calculated by 
\vspace{-1em}
\begin{equation}
\mathbf{x}_d(\mathbf{z}_q, \mathbf{p}_q, \mathbf{x}) = \sum_{m=1}^{M} \mathbf{W}_m A_{mq},
\label{eq:single_deform_attn_fun}
\vspace{-1 em}
\end{equation}
where
\vspace{-0.5 em}
\begin{equation}
   A_{mq} = \sum_{k=1}^{K} A_{mqk} \cdot \mathbf{W}'_m \mathbf{x}(\mathbf{p}_q + \Delta \mathbf{p}_{mqk}), 
\label{eq:attention_score}
\vspace{-0.5 em}
\end{equation}
$m$ indexes the attention head, $k$ indexes the sampled keys, and $K$ is the total number of sampled keys. $\mathbf{W}$ and $\mathbf{W}'$ are learnable weights. $\Delta \mathbf{p}_{mqk}$ and $A_{mqk}$ denote the sampling offset and attention weight of the $k$-th sampling point in the $m$-th attention head, where the sampling points are inferred by a linear layer. 

Most of recent keypoint detection and description frameworks~\cite{edstedt2024dedode, potje2024cvpr, Zhao2022ALIKE, zhao2023aliked} benefit from the use of multi-scale feature maps to capture fine details and broader spatial context. Deformable attention naturally extends to multi-scale feature maps~\cite{zhu2020deformable}, enabling the network to effectively adapt to features at different scales. Similar to~\cref{eq:single_deform_attn_fun}, multi-scale deformable attention feature $\mathbf{x}_{msd}$ can be calculated similar to~\cref{eq:single_deform_attn_fun} and~\cref{eq:attention_score}
\vspace{-1em}
\begin{equation}
\mathbf{x}_{msd}(\mathbf{z}_q, \mathbf{p}_q, \{\mathbf{x}^l\}_{l=1}^{L}) = \sum_{m=1}^{M} \mathbf{W}_m A'_{mq},
\label{eq:multi_deform_attn_fun}
\vspace{-1em}
\end{equation}
where
\begin{equation}
   A'_{mq} = \big[\sum_{l=1}^{L}\sum_{k=1}^{K} A_{mlqk} \cdot \mathbf{W}'_m \mathbf{x}_l(f_s(\hat{p_q}) + \Delta \mathbf{p}_{mlqk})\big].
\label{eq:attention_score}
\end{equation}

Here, $\hat{p_q} \in [0,1]^2$ is the normalized coordinates and function $f_s$ re-scales it to match the scale of input feature map of the $l$-th level.

\subsection{Network Architecture}
\label{sec:architecture}

\paragraph{Descriptor Branch}
The descriptor branch $\mathcal{F}_D$ extracts a dense feature map $D \in \mathbb{R}^{H/4 \times W/4 \times 256}$  and a matchability map $M$, which models that probability of a given feature vector can be matched.  


Using a ResNet-50~\cite{he2015deepresiduallearningimage} as the CNN backbone, $\mathcal{F}_{res}$ extracts feature maps at 4 scales, each with resolutions of $1/4, 1/8, 1/16, 1/32$ of the original image. An additional feature map at scale of $1/64$ is then included by applying a CNN on the last feature map. These features are positionally encoded and fed into $\mathcal{F}_e$ with multi-scale deformable attention~\cite{zhu2020deformable}, enabling the network to dynamically attend to relevant spatial locations across multiple scales. We use 4 encoder layers each with 8 attention heads and each head samples 8 points. The entire process of the Descriptor Branch is illustrated in~\cref{fig:pipeline}.
\vspace{-1em}
\paragraph{Keypoint Branch}\vspace{-0.5em} The keypoint branch is dedicated solely to detecting accurate keypoints without being influenced by descriptor estimation. $\mathcal{F}_{cnn}$ extracts feature maps at 4 different resolutions, each with a resolution of $1/1, 1/2, 1/8,$ and $1/32$ of the original image and feature dimension of 32. These features are subsequently upsampled to match the original image resolution and concatenated to produce a feature map of size $H \times W \times 128$. Then $\mathcal{F}_s$ is used to estimate the scoremap. DKD~\cite{Zhao2022ALIKE} is applied on the scoremap to perform non-maximum suppression and detect the sub-pixel-accurate keypoints. ~\cref{fig:pipeline} depicts the entire process of the Keypoint Branch.
 
\subsection{Feature Matching}
\label{sec:matching}
In this section, we describe how sparse matches and dense matches are obtained for a given image pair $I^1$ and $I^2$.
Our network outputs sparse keypoints $p_s^{1,2}$, descriptor maps $D^{1,2}$, and matchability maps $M^{1,2}$ all in a single forward pass. We bilinearly upsample $D^{1,2}$ to match the original resolution of $I^1$ and $I^2$. Descriptors $d^{1,2}$ are then sampled from $D^{1,2}$ using $p_s^{1,2}$. 

\paragraph{Sparse Matching}
\vspace{-0.5em}
\label{sparse}
Dual-Softmax operator~\cite{rocco2018neighbourhoodconsensusnetworks,tyszkiewicz2020disk} is used to establish correspondences between two descriptors $d^{1,2}$. The score matrix $S$ between the descriptors is first calculated by $S\left(i, j\right) = \frac{1}{\tau} \cdot \langle\ d^1 , d^2 \rangle$, where $\tau$ is the temperature. We can now apply softmax on both dimensions of $S$ to obtain the probability of soft mutual nearest neighbor (MNN) matching. 
The matching probability matrix $P$ is obtained by:
\begin{equation}
    P(i, j) = \operatorname{softmax}\left(S\left(i, \cdot \right)\right)_j \cdot \operatorname{softmax}\left(S\left(\cdot, j\right)\right)_i.
\label{eq:dual-softmax}
\end{equation}
Based on the confidence matrix $P$, 
we select matches $m_s$ with confidence higher than a threshold(0.01) to further enforce the mutual nearest neighbor criteria, which filters possible outlier matches.
\vspace{-1 em}
\paragraph{Semi-Dense Matching}
Recent works~\cite{sun2021loftr, chen2022aspanformer, Huang2023adamatcher, wang2022matchformer} demonstrated the benefits of semi-dense feature matching by improving coverage and number of inliers. We propose a novel dense matching module that produces accurate and geometry consistent dense matches.

Given images $I^1$ and $I^2$, our method can control the memory and computation usage by only selecting the top-K coarse keypoints $p_{c}^{1,2}$ according to their matchability score $M^{1,2}$. Then we can obtain the coarse matches $m_c$ using equation~\cref{eq:dual-softmax}, similar to how we obtain the sparse matches $m_s$. These matches are naturally coarse matches and require refinement for sub-pixel level accuracy as $M$'s resolution is only $1/4$ of the original resolution. Unlike existing methods~\cite{sun2021loftr, chen2022aspanformer} that crop fine level features around coarse matches in $I^1$ and $I^2$ and then predict offset in $I^2$, our work proposes a simple, efficient, and accurate module for semi-dense feature matching, inspired by~\cite{chen2024geometryawarefeaturematchinglargescale} that utilizes sparse correspondences in guiding semi-dense feature matching. 

Differing from~\cite{chen2024geometryawarefeaturematchinglargescale}, which iteratively reassigns the coarse matches of detector-free methods, we adopt a similar idea to refine coarse matches to fine matches. With $m_s$ obtained from sparse matching, we use the eight-point method to estimate a fundamental matrix $F$ that relates corresponding points between $I^1$ and $I^2$. We keep $p_{c}^1$ unchanged and refine $p_{c}^2$ by solving offsets ($\Delta x, \Delta y$) that satisfy the epipolar constraint defined by $F$. First, we convert $p_{c}^{1}$ to homogeneous coordinates $p_{h}^{1}$. Then, epipolar lines $L$ in $I^2$ for $p_h^1$ is computed by
\begin{equation}
    L = {F}\cdot{p_h^1},
\label{eq:epipolar}
\end{equation}
where each line is represented by coefficients $a, b$ and $c$. Offsets ($\Delta x, \Delta y$) are then calculated for each point in $p_c^2$ by solving the linear equation
\begin{equation}
    a\cdot(x + \Delta x) + b\cdot(y + \Delta y) + c = 0
\end{equation}
to enforce the epipolar constraint. Then we can get offsets
\begin{equation}
    \Delta x = \frac{b(bx - ay) - ac}{a^2 + b^2 + \epsilon} - x, \quad \Delta y = \frac{a(ay - bx) - bc}{a^2 + b^2 + \epsilon} - y,
\end{equation}
where $(x, y)$ are the pixel coordinates of $p_c^2$. We finally filter out points with offsets that are larger than the patch size (4) because they are moved outside the matched patch and likely outliers. We then obtain refined dense matches $m_c$ by $(x + \Delta x, y + \Delta y)$. In~\cref{sec:ablation}, we show that without this refinement module, the performance of dense matching degrades significantly.

\subsection{Implementation Details}
\label{sec:implementation}
We train \shortname with pseudo ground truth correspondences from the MegaDepth dataset~\cite{li2018megadepth} and our Air-to-Ground dataset. We collect data from Internet drone videos and ground images of different famous landmarks, more details can be found in the supplementary. Following~\cite{sun2021loftr, sarlin20superglue}, we use poses and depths to establish pixel correspondences. For every given image pair $(I^1, I^2)$ with $N$ matching pixels, $m_{gt} \in R^{N\times4}$ is defined such that the first two columns are the pixel coordinates of points in $I^1$ and the last two columns are those in $I^2$ obtained through a warping function $f_{warp}$. We train the descriptor branch and the keypoint branch separately.

\paragraph{Training Descriptor Branch} 
\vspace{-1 em}
To supervise the local feature descriptors, we apply focal loss~\cite{lin2018focallossdenseobject} to focus on challenging examples and reduce the influence of easily classified matches. Given descriptor maps \( D^{1} \) and \( D^{2} \), we sample descriptor sets \( d_{gt}^{1} \) and \( d_{gt}^{2} \) based on the ground truth matchability \( m_{gt} / k\), where \( k \) is the patch size. The \( i \)-th rows, \( d_{gt}^1(i) \) and \( d_{gt}^2(i) \), represent descriptors corresponding to the same points in the images \( I^1 \) and \( I^2 \), respectively. Using these descriptors, we compute a probability matrix \( P \) as described in~\cref{eq:dual-softmax}. Our supervision targets positive correspondences only, represented by the diagonal elements \( P_{(i, i)} \) of \( P \). By enforcing \( P_{(i, i)} \) to be close to 1, we calculate the focal loss \( \mathcal{L}_{focal} \) as follows:
\begin{equation}
    \mathcal{L}_{focal} = -\alpha \cdot (1 - P_{(i, i)})^{\gamma} \cdot \log(P_{(i, i)}),
\label{eq:focal}
\end{equation}
where \( \alpha = 0.25 \) and \( \gamma = 2 \).

\begin{figure*}[!ht]
    \vspace{-1 em}
    \centering
    \includegraphics[width=0.95\linewidth]{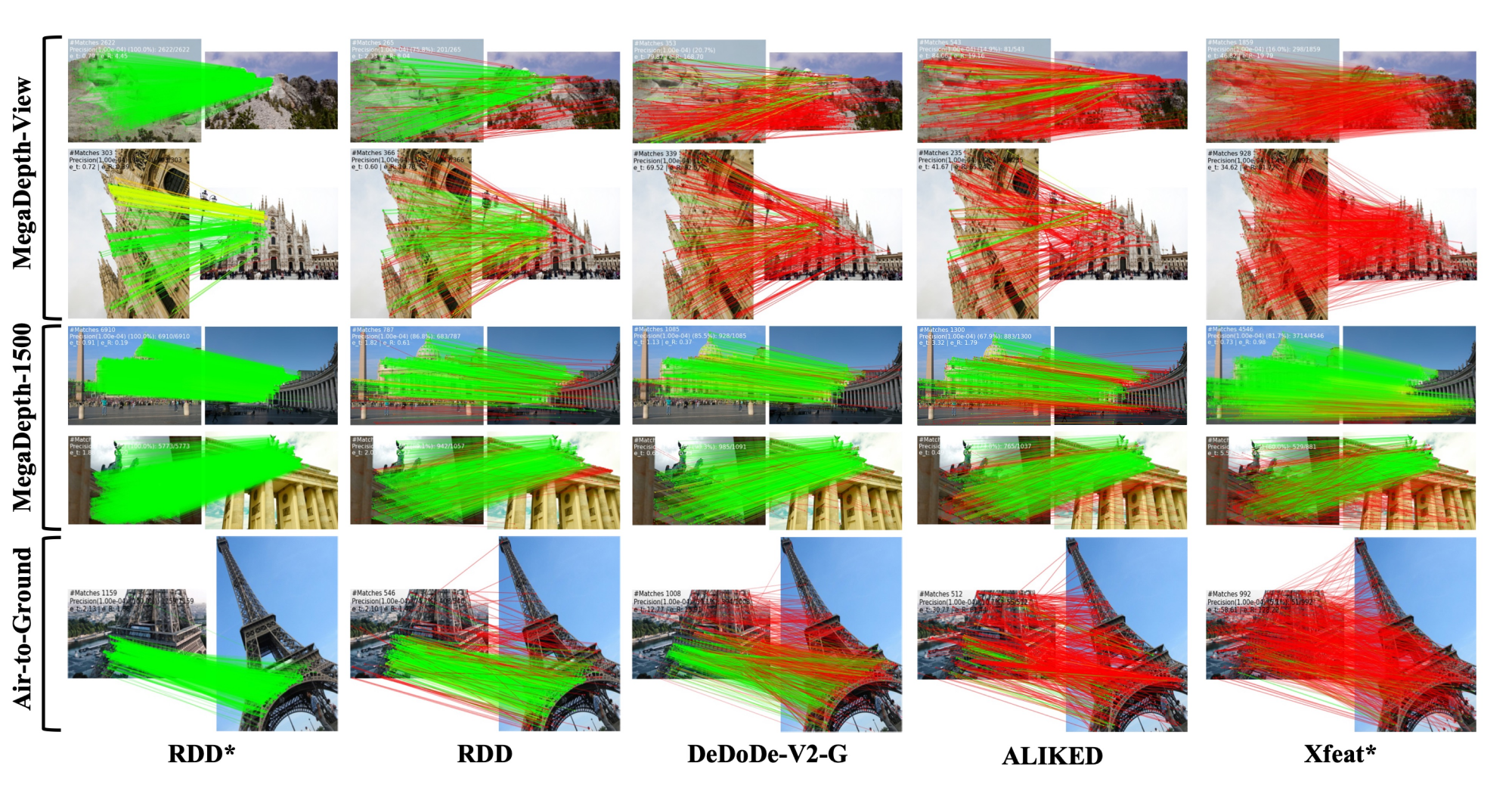}
    \vspace{-0.8 em}
    \caption{\textbf{Qualitative Results on MegaDepth.} RDD* and \shortname are qualitatively compared to DeDoDe-V2-G~\cite{edstedt2024dedode,edstedt2024dedodev2} ALIKED~\cite{zhao2023aliked} and XFeat*~\cite{potje2024cvpr}. RDD* and \shortname are more robust compared DeDoDe-G* and ALIKED under challenging scenarios like large scale and viewpoint changes. The red color indicates epipolar error beyond $1\times10^{-4}$ (in the normalized image coordinates)}
    \label{fig:qualitative}
    \vspace{-1em}
\end{figure*}

To supervise the matchability map, we apply a modified focal loss that incorporates binary cross-entropy (BCE) on the matchability map \( M^{1,2} \). The ground truth matchability map \( M_{gt}^{1,2} \) is generated using \( m_{gt} / k \). We first compute the binary cross-entropy loss \( \mathcal{L}_{BCE} \) as
\begin{equation}\resizebox{0.85\linewidth}{!}{%
    $\mathcal{L}_{BCE} = -\left( M_{gt} \cdot \log(M) + (1 - M_{gt}) \cdot \log(1 - M) \right),$%
    }
\end{equation}
and then replace \( P_{(i, i)} \) in~\cref{eq:focal} with \(\lambda = e^{-\mathcal{L}_{BCE}} \), resulting in the matchability loss \( \mathcal{L}_{matchability} \):
\begin{equation}
    \mathcal{L}_{matchability} = -\alpha \cdot (1 - \lambda)^{\gamma} \cdot \mathcal{L}_{BCE}.
\label{eq:match}
\end{equation}

The final loss for the descriptor branch is defined as $\mathcal{L}_D = \mathcal{L}_{focal} + \mathcal{L}_{matchability}$. We train the descriptor branch on the mixture of MegaDepth dataset~\cite{li2018megadepth} and collected Air-to-Ground dataset, with images resized to 800. The training uses a batch size of 32 image pairs and the AdamW optimizer~\cite{loshchilov2019decoupledweightdecayregularization}. We use a base learning rate of \( 1 \cdot 10^{-4} \) with a step learning rate scheduler that updates the learning rate every 2000 steps after 20000 steps with \( \gamma = 0.5 \). The descriptor branch converges after 1 day of training on 8 NVIDIA H100 GPU.

\paragraph{Training Keypoint Branch}\vspace{-1 em}
We freeze the weights of descriptor branch once it converges and only supervise the keypoint branch to detect accurate and repeatable keypoints. We apply reprojection loss $\mathcal{L}_{reprojection}$, reliability loss $\mathcal{L}_{reliability}$ similar to~\cite{Zhao2022ALIKE, dusmanu2019d2net, luo2020aslfeat} and dispersity peaky loss $\mathcal{L}_{peaky}$~\cite{Zhao2022ALIKE} to train repeatable and reliable keypoints.

Given detected keypoint $p_s^1$ from $I^1$ and $p_s^2$ from $I^2$. We obtain the warped keypoints $p_s^{1 \to 2}$ using $f_{warp}$ and we find their closest keypoints from $p_s^2$ with distance less than 5 pixels as matched keypoints. Then, $dist_{1 \to 2}$ is the reprojection distance between the matched $p_s^{1}$ and $p_s^{2}$. We define the reprojection loss in a symmetric way:
\vspace{-0.4 em}
\begin{equation}
    \mathcal{L}_{reprojection} = \frac{1}{2}(dist_{1 \to 2} + dist_{2 \to 1}),
\label{eq:reproject}
\end{equation}
For keypoint \( p_s^1 \) and its warped keypoint \( p_s^{1 \to 2} \) with scores \( s^1 \) and \( s^{1 \to 2} \), and the probability map \( P \) as described in~\cref{sparse}, the reliability map is defined as
\begin{equation}
    R = \exp\left(\frac{P - 1}{t_{rel}}\right),
\end{equation}
where \( t_{rel} \) is the temperature parameter. Considering all valid \( p_s^1 \) and warped points \( p_s^{1 \to 2} \), we sample the reliability \( r^1 \) from \( R \). The reliability loss for \( p_s^1 \) is then defined as
\begin{equation}
    \mathcal{L}_{\text{reliability}}^1 = \frac{1}{N_1} \sum_{p_s^1,p_s^{1 \to 2}} \frac{s^1 \cdot s^{1 \to 2}}{\sum_{p_s^1,p_s^{1 \to 2}} s^1 \cdot s^{1 \to 2}} (1 - r^1).
\end{equation}
Similarly, the total reliability loss is defined symmetrically as
\begin{equation}
    \mathcal{L}_{\text{reliability}} = \frac{1}{2} \left(\mathcal{L}_{\text{reliability}}^1 + \mathcal{L}_{\text{reliability}}^2\right).
\end{equation}

~\cref{eq:reproject} optimizes the scores of keypoints in local window through the soft term $[\hat i, \hat j]_{soft}^T$ of~\cref{eq:dkd}, which might affect the first term $[x',y']_{nms}^T$ of the equation. In order to align their optimization direction, we define the dispersity peak loss similiar to~\cite{Zhao2022ALIKE} as
\begin{equation}
	\mathcal{L}_{peaky} = \frac{1}{N^2} \sum_{0 \le i,j<N} d(i,j) s'(i,j),
\end{equation}
\vspace{-1 em}

where
\begin{equation}
	d(i,j) = \left \{ ||{[i,j] - [\hat i, \hat j]_{soft}}||_p \ | \ 0\le i,j < N \right \},
\end{equation}
$N$ is the window size and $s'(i, j)$ are softmax scores defined in~\cref{eq:softmax}.

The overall loss for the keypoint branch is defined as $\mathcal{L}_K = \mathcal{L}_{reprojection} + \mathcal{L}_{reliability}+\mathcal{L}_{peaky}$. We use the DKD~\cite{Zhao2022ALIKE} with a window size of $N=5$ to detect the top 500 keypoints, and we randomly sample 500 keypoints from non-salient positions. We train the keypoint branch similarly to the descriptor branch, the keypoint branch converges after 4 hours of training on an NVIDIA H100 GPU.
\vspace{-1 em}
\section{Experiments}
\label{sec:experiments}
In this section, we demonstrate the performance of our method in Relative Pose Estimation on public benchmark datasets MegaDepth~\cite{li2018megadepth} and collected Air-to-Ground dataset, and in Visual Localization using Aachen Day-Night dataset~\cite{sattler2018benchmarking6dofoutdoorvisual}. All experiments are conducted on an NVIDIA H100 GPU. 

\subsection{Relative Pose Estimation}\label{sub:Pose}

\paragraph{Dataset} Following~\cite{edstedt2024dedode, potje2024cvpr, dusmanu2019d2net}, we use the standard MegaDepth-1500 benchmark from D2-Net~\cite{dusmanu2019d2net} for outdoor pose estimation and adopt the same test split. To further validate the performance of \shortname in challenging scenarios, we construct two benchmark datasets. MegaDepth-View is derived from MegaDepth~\cite{li2018megadepth} test scenes, focusing on pairs with large viewpoint and scale changes; additionally, we sampled 1500 pairs from our collected Air-to-Ground dataset. This benchmark is designed to validate feature matching performance for large-baseline outdoor cross-view images, which is naturally challenging because of geometric distortion and significant viewpoint changes. 

\begin{table}
\small
 \centering
 \caption{\textbf{SotA comparison on the MegaDepth~\cite{li2018megadepth}}. Results are measured
 in AUC (higher is better). Top 4,096 features used to all sparse matching methods. Best in bold, second best underlined.}\vspace{-0.1em}
 \resizebox{0.99\columnwidth}{!}{%
 \begin{tabular}{lcccccc}
    \toprule      
    \multirow{3}{*}{\textbf{Method}} &\multicolumn{3}{c}{\textbf{MegaDepth-1500}} & \multicolumn{3}{c}{\textbf{MegaDepth-View}}  \\ 
                                     & \multicolumn{3}{c}{AUC}    &   \multicolumn{3}{c}{AUC}    \\
                                     & @5\degree & @10\degree & @20\degree    &      @5\degree & @10\degree & @20\degree      \\
    \midrule
    \textbf{\emph{Dense}} & & & & & & \\
    DKM~\cite{edstedt2023dkm}~\tiny{CVPR'23}  &  60.4 & 74.9 & 85.1 & 67.4 & 80.0 & 88.2 \\
    RoMa~\cite{edstedt2024roma}~\tiny{CVPR'24} &  62.6 & 76.7 & 86.3 & 69.9 & 81.8 & 89.4 \\
    \midrule
    \textbf{\emph{Semi-Dense}} & & & & & & \\
    LoFTR~\cite{sun2021loftr}~\tiny{CVPR'21}& 52.8 & 69.2 & 81.2 & 50.6 & 65.8 & 77.9 \\
    ASpanFormer~\cite{chen2022aspanformer}~\tiny{ECCV'22} & 55.3  & 71.5 & 83.1 & 61.3 & 75.2 & 84.6\\
    ELoFTR~\cite{wang2024eloftr}~\tiny{CVPR'24} & 56.4 & 72.2 & 83.5 & 60.2 & 74.8 & 84.7 \\
    XFeat*~\cite{potje2024cvpr}~\tiny{CVPR'24} & 38.6 & 56.1 & 70.5 & 23.9 & 37.7 & 51.8\\
    RDD* & 51.3 & 67.1 & 79.3 & 41.5 & 55.7 & 66.7 \\
    \midrule
    \textbf{\emph{Sparse with Learned Matcher}} & & & & & & \\
    SP~\cite{superpoint2018}+SG~\cite{sarlin20superglue}~\tiny{CVPR'19} & 49.7 & \underline{67.1} & \underline{80.6} & 51.5 & 65.5 & 76.9 \\      
    SP~\cite{superpoint2018}+LG~\cite{lindenberger2023lightglue}~\tiny{ICCV'23} & \underline{49.9} & 67.0 & 80.1 & \underline{52.4} & \underline{67.3} & \underline{78.5} \\
    Dedode-V2-G~\cite{edstedt2024dedode,edstedt2024dedodev2}+LG~\cite{lindenberger2023lightglue}~\tiny{ICCV'23} &  44.1 & 62.1 & 76.5 & 41.9 & 57.1 & 70.0 \\
    RDD+LG~\cite{lindenberger2023lightglue}~\tiny{ICCV'23} &  \textbf{52.3} & \textbf{68.9} & \textbf{81.8} & \textbf{54.2} & \textbf{69.3} & \textbf{80.3} \\
    \midrule
    \textbf{\emph{Sparse with MNN}} & & & & & & \\
    SuperPoint~\cite{superpoint2018}~\tiny{CVPRW'18} & 24.1 & 40.0 & 54.7 & 7.50 & 13.3 & 21.1 \\
    DISK~\cite{tyszkiewicz2020disk}~\tiny{NeurIps'20} & 38.5 & 53.7 & 66.6 & 30.4 & 41.9 & 51.6 \\
    ALIKED~\cite{zhao2023aliked}~\tiny{TIM'23} & 41.8 & 56.8 & 69.6 & 30.0 & 41.8 & 53.2 \\
    XFeat~\cite{potje2024cvpr}~\tiny{CVPR'24} & 24.0 & 40.1 & 55.8 & 8.82 & 16.5 & 26.8 \\
    DeDoDe-V2-G~\cite{edstedt2024dedode,edstedt2024dedodev2}~\tiny{CVPRW'24, 3DV'24} & \underline{47.2} & \underline{63.9} & \underline{77.5} & \underline{33.1} & \underline{47.6} & \underline{60.2}\\
    RDD & \textbf{48.2} & \textbf{65.2} & \textbf{78.3} & \textbf{38.3} & \textbf{53.1} & \textbf{65.6} \\
 \bottomrule
 \end{tabular}
 }
 \vspace{-1 em}
 \label{tab:megadepth}
\end{table}

\begin{table}
 \small
     \centering
     \caption{\textbf{SotA comparison on proposed Air-to-Ground benchmark}. Keypoints and descriptors are matched using dual-softmax MNN. Measured in AUC (higher is better). Best in bold, second best underlined.}
     \resizebox{0.9\columnwidth}{!}{%
     \begin{tabular}{l lll}
     \toprule
      Method & $@5^{\circ}$&$@10^{\circ}$&$@20^{\circ}$\\
        \midrule
        SuperPoint~\cite{superpoint2018}~\tiny{CVPRW'18} & 1.89 & 3.56 & 6.81 \\
        DISK~\cite{tyszkiewicz2020disk}~\tiny{NeurIps'20} & 19.2 & 27.1 & 34.6 \\
        ALIKED~\cite{zhao2023aliked}~\tiny{TIM'23} & 12.0 & 17.8 & 25.8 \\
        XFeat~\cite{potje2024cvpr}~\tiny{CVPR'24} & 6.12 & 11.4 & 17.5 \\
        DeDoDe-V2-G~\cite{edstedt2024dedode,edstedt2024dedodev2}~\tiny{CVPRW'24,3DV'24} & \underline{31.5} & \underline{45.3} & \underline{58.3} \\
        \shortname & \textbf{41.4} & \textbf{56.0} & \textbf{67.8} \\

     \bottomrule
     \end{tabular}
     }
     \vspace{-1 em}
     \label{tab:air-ground}
 \end{table}

\paragraph{Metrics and Comparing Methods}
\vspace{-0.7em}
We report the AUC of recovered pose under threshold of (5\degree, 10\degree\ ,and 20\degree). We use RANSAC to estimate the essential matrix. We compared \shortname against the state-of-the-art detector/descriptor methods~\cite{superpoint2018,tyszkiewicz2020disk,Zhao2022ALIKE,zhao2023aliked,gleize2023silksimplelearned,edstedt2024dedode, potje2024cvpr} and learned matching methods~\cite{sarlin20superglue,lindenberger2023lightglue,sun2021loftr,chen2022aspanformer,wang2024eloftr,edstedt2023dkm,edstedt2024roma} on MegaDepth~\cite{li2018megadepth} dataset. For Air-to-Ground dataset, \shortname is compared against detector/descriptor methods. More results on Air-to-Ground data can be found in the supplementary. For all sparse methods, we use resolution whose larger dimension is set to 1,600 pixels and use top 4,096 features, while all other learned matching methods follow experiment settings mentioned in their paper.
\vspace{-0.5 em}
\paragraph{Results} \vspace{-0.7em}The results are presented in~\cref{tab:megadepth} and~\cref{tab:air-ground}, and visually in~\cref{fig:qualitative}. RDD demonstrates superior performance in the sparse matching setting compared to the state-of-the-art method DeDoDe-V2-G~\cite{edstedt2024dedode, edstedt2024dedodev2}, especially in more challenging scenarios. RDD with learned matcher LightGlue~\cite{lindenberger2023lightglue} also outperforms previous methods.

The results on MegaDepth-View and Air-to-Ground prove the effectiveness of our method under challenging scenarios with significant improvements over previous methods. \shortname outperforms all previous feature detector/descriptor methods under large viewpoints and scale changes. 


\subsection{Homography Estimation}\label{sub:homo}

\paragraph{Datasets} We evaluate the quality of correspondences estimated by \shortname on HPatches dataset~\cite{hpatches_2017_cvpr} for Homography Estimation. HPatches contains 52 sequences under significant illumination changes and 56 sequences that exhibit large variation in viewpoints. We use RANSAC to robustly estimate the homography given the estimated correspondences. All images are resized such that their shorter size is equal to 480 pixels.  

\begin{table}
	\centering
	\caption{{\bf Homography estimation on HPatches.} All methods perform well for illumination sequences. \shortname provides high quality homography estimation especially when there are significant viewpoint changes. Best in bold, second best underlined.}
    \resizebox{0.99\columnwidth}{!}{%
	\begin{tabular}{lcccccc}
		\toprule      
		\multirow{3}{*}{\textbf{Method}} &\multicolumn{3}{c}{\textbf{Illumination}} & \multicolumn{3}{c}{\textbf{Viewpoint}}  \\ 
         & \multicolumn{3}{c}{MHA}    &   \multicolumn{3}{c}{MHA}    \\
         & @3px & @5px & @10px     &      @3px & @5px & @10px  \\
		\midrule
        SuperPoint~\cite{superpoint2018}~\tiny{CVPRW'18} & 93.0 & 98.0 & \textbf{99.9} & 70.0 & 82.0 & 87.0 \\
        DISK~\cite{tyszkiewicz2020disk}~\tiny{NeurIps'20} & 96.0 & 98.0 & 98.0 & 72.0 & 81.0 & 84.0 \\
        ALIKED~\cite{zhao2023aliked}~\tiny{TIM'23} & \textbf{97.0} & \textbf{99.0} & 99.0 & \textbf{78.0} & \underline{85.0} & \underline{88.0} \\
        DeDoDe-G~\cite{edstedt2024dedode}~\tiny{3DV'24} & \underline{96.0} & \textbf{99.0} & \textbf{99.9} & 68.0 & 77.0 & 80.0 \\
        XFeat~\cite{potje2024cvpr}~\tiny{CVPR'24}     & 90.0 & 96.0 & 98.0 & 55.0 & 72.0 & 80.0 \\
        \shortname & 93.0 & \textbf{99.0} & \textbf{99.9} & \underline{76.0} & \textbf{86.0} & \textbf{90.0} \\
		\bottomrule		
	\end{tabular}  
    }
	\label{tab:homography}
    \vspace{-1 em}
\end{table}
\vspace{-1 em}
\paragraph{Metrics and Comparing Methods} We follow~\cite{potje2024cvpr, Zhao2022ALIKE} to estimate the mean homogeneity accuracy (MHA) with a predefined threshold of \{3, 5, 7\} pixels. Accuracy is computed using the average corner error in pixels by warping reference image corners onto target images using both ground truth and estimated homographies.

\vspace{-1 em}
\paragraph{Results on HPatches} In Table~\ref{tab:homography}, \shortname shows a similar performance in illumination sequences compared to previous accurate keypoint detectors and descriptors, while outperforming most of the other methods on viewpoint changes, reinforcing the robustness of our proposed method in handling challenging cases such as large baseline pairs. 

\subsection{Visual Localization}

\paragraph{Dataset} We further validate the performance of~\shortname on the task of visual localization, which estimates the poses of a given query image with respect to the corresponding 3D scene reconstruction. We evaluate~\shortname on the Aachen Day-Night dataset~\cite{sattler2018benchmarking6dofoutdoorvisual}. It focuses on localizing high-quality night-time images against a day-time 3D model. There are 14,607 images with changing conditions of weather, season, viewpoints, and day-night cycles.

\begin{table}[!ht]

    \centering
    \caption{{\bf Visual Localization on Aachen day-night.} \shortname outperforms previous state-of-the-art methods with MNN matching for more challenging night setting. Best in bold, second best underlined.}
    \resizebox{0.85\columnwidth}{!}{%
    \begin{tabular}{lcccc}
        \toprule
        \multirow{2}{1.3cm}[-.4em]{Methods}
        & Day & & Night\\
        \cmidrule(lr){2-4}
        & \multicolumn{3}{c}{(0.25m,2°) / (0.5m,5°) / (1.0m,10°)}\\
        \midrule
        SuperPoint~\cite{superpoint2018}~\tiny{CVPRW'18}     & \textbf{87.4} / 93.2 / \underline{97.0} && 77.6 / 85.7 / 95.9 \\
        DISK~\cite{tyszkiewicz2020disk}~\tiny{NeurIps'20}            & 86.9 / \textbf{95.1} / \textbf{97.8} && 83.7 / 89.8 / \textbf{99.0} \\
        ALIKE~\cite{Zhao2022ALIKE}~\tiny{TM'22}            & 85.7 / 92.4 / 96.7 && 81.6 / 88.8 / \textbf{99.0} \\
        ALIKED~\cite{zhao2023aliked}~\tiny{TIM'23}           & 86.5 / 93.4 / 96.8 && \underline{85.7} / \underline{91.8} / 96.9 \\
        XFeat~\cite{potje2024cvpr}~\tiny{CVPR'24}           & 84.7  / 91.5 / 96.5 && 77.6 / 89.8 / \underline{98.0} \\
        \shortname & \underline{87.0}  / \underline{94.2} / \textbf{97.8} && \textbf{86.7} / \textbf{92.9} / \textbf{99.0} \\
        \bottomrule
    \end{tabular}
    }
    \label{tab:aachen}
\end{table}
\vspace{-1.5 em}

\begin{table}[!ht]
    \centering
    \caption{\textbf{Ablation study on MegaDepth.} We ablate the design choices for architecture and training strategies for relative pose estimation on MegaDepth.}
    \resizebox{0.8\columnwidth}{!}{%
    \begin{tabular}{lcc}  
    \toprule
    \textbf{ Variant } & \multicolumn{2}{c} {\textbf{AUC@$5^\circ$}}  \\
    & \shortname & RDD* \\
    \midrule
    Full                                  & \textbf{48.2}  & \textbf{51.3} \\
    Larger patch size $s=8$               & 44.6 & 49.4 \\
    Less sample points $N_{p_q} = 4$      & 46.5 & 49.9 \\
    No keypoint branch                    & 44.1 & -    \\
    Joint training of two branches        & 42.8 & 46.9 \\
    No match refinement                   & -    & 41.3 \\
    Without Air-to-Ground Data            & 47.4 & 50.4 \\
    \bottomrule
    \end{tabular}
    }
    \label{tab:ablation}
    \vspace{-2 em}
\end{table}

\paragraph{Metrics and Comparing Methods} Following~\cite{potje2024cvpr}, HLoc~\cite{sarlin2019coarse} is used to evaluate all approaches~\cite{potje2024cvpr, Zhao2022ALIKE, superpoint2018,sarlin20superglue,tyszkiewicz2020disk,zhao2023aliked}. We resize the images such that their longer side is equal to 1,024 pixels and sample the top 4,096 keypoints for all approaches. We use the evaluation tool provide by~\cite{sattler2018benchmarking6dofoutdoorvisual} to calculate the metrics, the AUC of estimated camera poses under threshold of $\{0.25$m$, 0.5$m$, 5$m$\}$ for translation errors and $\{2\degree, 5\degree, 10\degree\}$ for rotation errors respectively.  

\paragraph{Results} ~\cref{tab:aachen} presents the results on visual localization. \shortname outperforms all approaches in the challenging night setting. These results further validate the effectiveness of \shortname. 


\subsection{Ablation}
\label{sec:ablation}
\vspace{-0.1em}
To better understand RDD, we evaluate 5 variants of RDD, with the results shown in ~\cref{tab:ablation}:
1) Increasing the patch size of the transformer encoder results in a significant drop in accuracy.
2) Reducing the number of sampled points per query in deformable attention degrades pose estimation accuracy.
3) Removing the keypoint branch and using upsampled matchibility map as keypoint score map with NMS negatively affect the performance.
4) Joint training of the descriptor and keypoint branches leads to a significant drop in performance, as the weights of both branches are updated when only one task fails, making it even harder for the descriptor branch to converge.
5) Simply concatenating the semi-dense matches from matchability results in worse performance.
6) Training without Air-to-Ground dataset slightly damages the performance.

\vspace{-0.5 em}
\section{Limitations and Conclusion}
We present RDD, a robust feature detection and description framework based on deformable transformer, excelling in sparse image matching and demonstrating strong resilience to geometric transformations and varying conditions. By leveraging deformable attention, RDD efficiently extracts geometric-invariant features, enhancing accuracy and adaptability compared to traditional methods. RDD lacks explicit data augmentation during training and depends on confident priors from sparse correspondences which might lead to the failure of semi-dense matching. Training with data augmentation and improving the refinement module with visual features might boost the performance. RDD marks a significant advancement in keypoint detection and description, paving the way for impactful applications in challenging 3D vision tasks like cross-view scene modeling and visual localization.

\vspace{-0.5 em}

\section{Acknowledgement}
Supported by the Intelligence Advanced Research Projects Activity (IARPA) via Department of Interior/ Interior Business Center (DOI/IBC) contract number 140D0423C0075. The U.S. Government is authorized to reproduce and distribute reprints for Governmental purposes notwithstanding any copyright annotation thereon. Disclaimer: The views and conclusions contained herein are those of the authors and should not be interpreted as necessarily
representing the official policies or endorsements, either expressed or implied, of IARPA, DOI/IBC, or the U.S. Government. We would like to thank Yayue Chen for her help with visualization.
\clearpage
\setcounter{page}{1}
\maketitlesupplementary

In this supplementary material, we present a detailed overview of our datasets and how the Air-to-Ground dataset is collected. We also provide more results on Air-to-Ground and an expanded set of qualitative results. Moreover, we evaluate the runtime performance of our method as well as the performance with better RANSAC solver.

\section{Datasets}
\label{sec:dataset}
To improve the robustness of \shortname in challenging scenarios, we proposed a training dataset Air-to-Ground and 2 benchmark datasets to better understand the performance.

\subsection{MegaDepth-View}
\label{sec:mega-view}
MegaDepth-View is derived from the test scenes of MegaDepth~\cite{li2018megadepth}. MegaDepth~\cite{li2018megadepth} is a large-scale outdoor dataset containing over 1 million Internet images from 196 different locations. Camera poses are reconstructed using COLMAP~\cite{schoenberger2016sfm, schoenberger2016mvs}, and depth maps are generated via multi-view stereo. The test scenes comprise 8 distinct locations around the world, ensuring diversity in the testing data.

\begin{figure}
    \centering
    \includegraphics[width=0.99\linewidth]{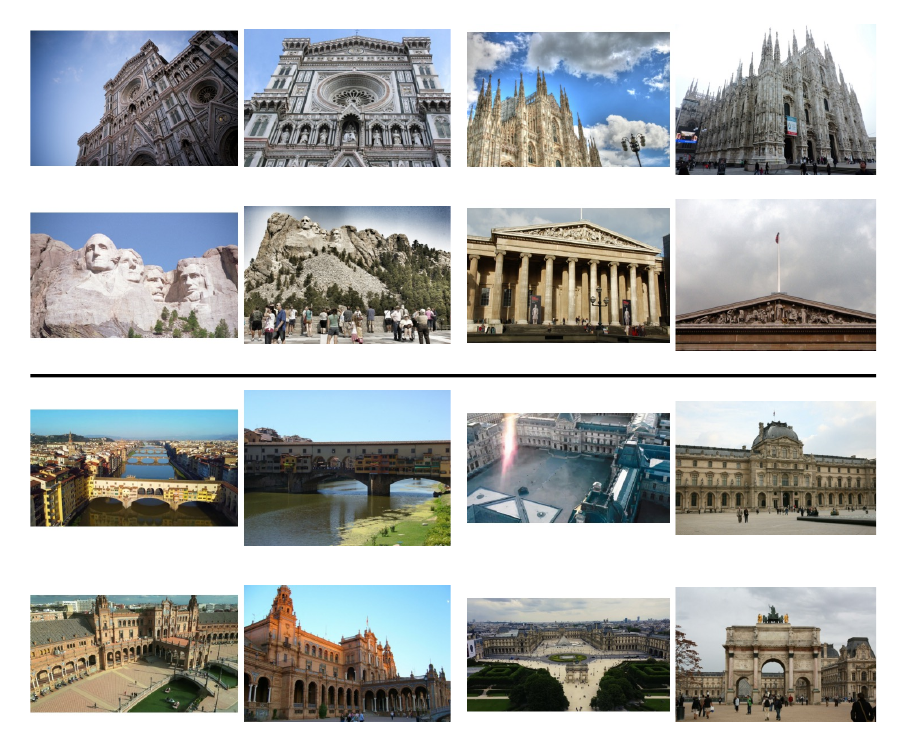}
    \caption{\textbf{Example Pairs from MegaDepth-View and Air-to-Ground} The top section shows example pairs from the MegaDepth-View benchmark, which emphasizes large viewpoint shifts and scale differences. The bottom section presents example pairs from the Air-to-Ground dataset/benchmark, designed for the novel task of matching aerial images with ground images.}
    \label{fig:examples}
\end{figure}

We focus on image pairs that exhibit significant viewpoint shifts and scale changes. For all possible image pairs, we first compute the overlap between two images by bi-directionally warping them using camera poses and depth. Then, we select image pairs with more than 2,000 matching pixels but fewer than 20,000 matching pixels. This process resulted in a total of 1,487 image pairs, forming our MegaDepth-View benchmark. Example pairs are shown in~\cref{fig:examples}.

\begin{figure*}
    \centering
    \includegraphics[width=0.99\linewidth]{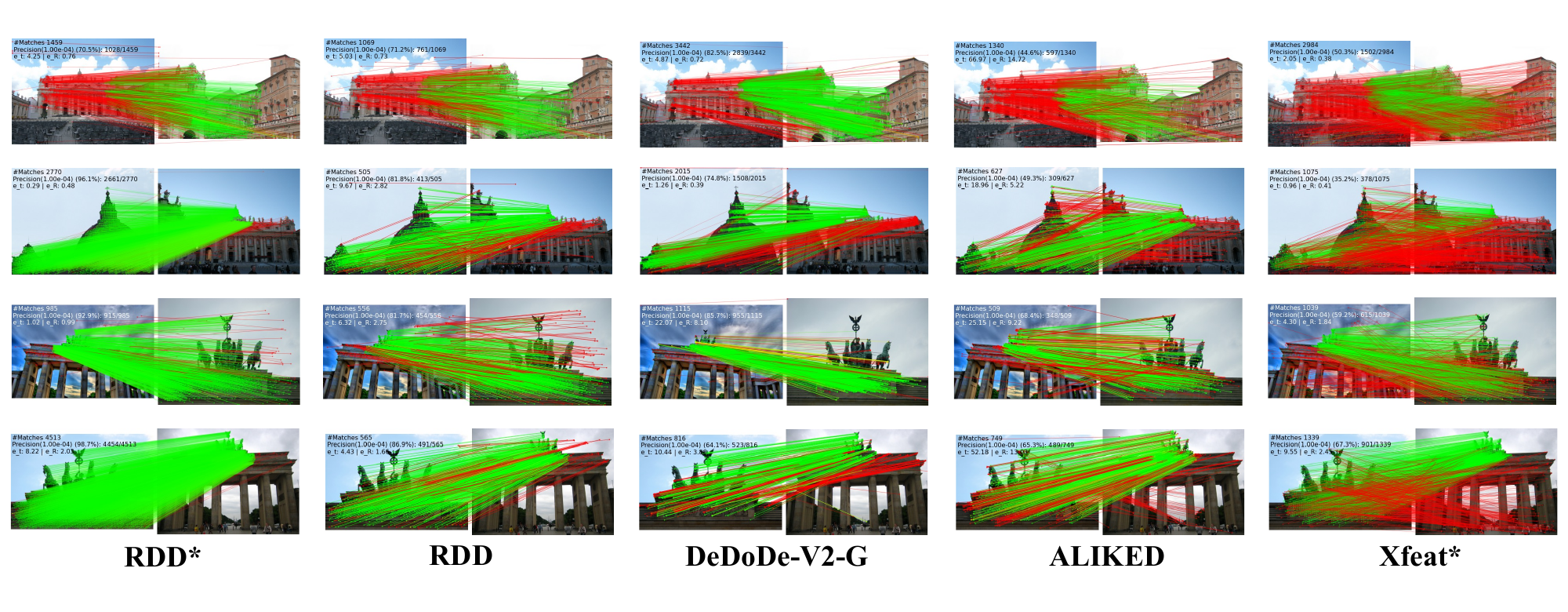}
    \caption{\textbf{More Qualitative Results on MegaDepth-1500.} RDD* outperforms DeDoDe-G* in semi-dense matching setting with 30,000 keypoints with a better runtime efficiency~\cref{tab:time_ex}. The red color indicates epipolar error beyond $1\times10^{-4}$ (in the normalized image coordinates).}
    \label{fig:mega1500-more}
\end{figure*}

\begin{figure*}
    \centering
    \includegraphics[width=0.975\linewidth]{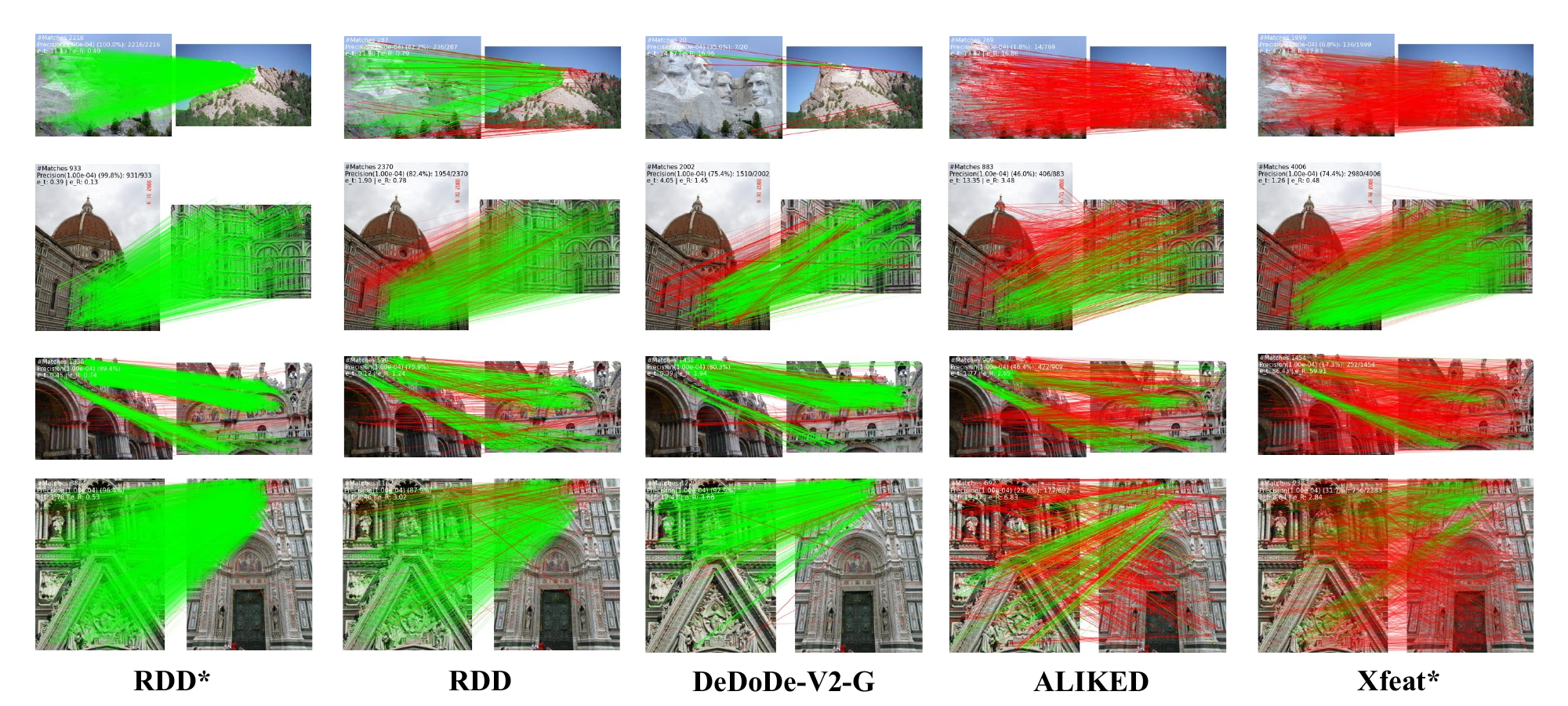}
    \caption{\textbf{More Qualitative Results on MegaDepth-View.} RDD and RDD* are robust under large viewpoint shifts and scale differences. The red color indicates epipolar error beyond $1\times10^{-4}$ (in the normalized image coordinates).}
    \label{fig:megaview-more}
\end{figure*}

\subsection{Air-to-Ground}
\label{sec:air}
\subsubsection{Data Collection}
3D reconstruction from imagery captured at multiple altitudes has increasingly garnered attention, driven by the growing UAV industry. Finding reliable correspondences between cross-view imagery has become a significant bottleneck in this domain~\cite{chen2024geometryawarefeaturematchinglargescale}. \shortname is designed to maintain robustness under large camera baselines and aims to enhance the accuracy and reliability of its downstream applications like 3D reconstruction for cross-view imagery. To validate RDD's ability to address such bottlenecks and evaluate its robustness, we collected the first large benchmark dataset focusing on cross-view imagery. This dataset includes a total of 41 famous locations around the world, such as the Eiffel Tower, Louvre Museum, Sacré-Cœur Basilica, London Bridge, Ponte Vecchio, Las Vegas Strip, Altare della Patria, Flatiron Building, Jackson Square, and Plaza de España. The dataset has around 27,000 images and over 600,000 air-to-ground image pairs.

Inspired by MegaDepth~\cite{li2018megadepth}, we use COLMAP~\cite{schoenberger2016sfm,schoenberger2016mvs} to reconstruct camera poses and estimate depth maps. Differing from MegaDepth~\cite{li2018megadepth} which uses internet images, we collect Internet drone videos and ground images. Drone videos allow us to track frames from the ground up to the air, generating one comprehensive 3D reconstruction including both ground images and frames extracted from drone videos.

\begin{table}
     \centering
     \caption{\textbf{More results on proposed Air-to-Ground benchmark}. Results are measured in AUC (higher is better). Best in bold, second best underlined.}
     \begin{tabular}{l lll}
     \toprule
      Method & $@5^{\circ}$&$@10^{\circ}$&$@20^{\circ}$\\
        \midrule
        \textbf{\emph{Dense}} & & & \\
        DKM~\cite{edstedt2023dkm}~\tiny{CVPR'23}  & 65.0  & 77.2 & 85.7 \\
    RoMa~\cite{edstedt2024roma}~\tiny{CVPR'24} & 71.3 & 82.4 & 89.5 \\
    \midrule
    \textbf{\emph{Semi-Dense}} & & & \\
    LoFTR~\cite{sun2021loftr}~\tiny{CVPR'21}& 21.5 & 33.8 & 45.6 \\
    ASpanFormer~\cite{chen2022aspanformer}~\tiny{ECCV'22} & 45.8 & 60.0 & 71.0 \\
    ELoFTR~\cite{wang2024eloftr}~\tiny{CVPR'24} & 49.4 & 62.8 & 73.2 \\

    XFeat*~\cite{potje2024cvpr}~\tiny{CVPR'24} & 12.0 & 19.2 & 27.9 \\

    RDD* & 43.8 & 55.3 & 64.9 \\
    \midrule
    \textbf{\emph{Sparse with Learned Matcher}} & & &  \\
    SP~\cite{superpoint2018}+SG~\cite{sarlin20superglue}~\tiny{CVPR'19} & 42.0 & 56.3 & 67.7 \\      
    SP~\cite{superpoint2018}+LG~\cite{lindenberger2023lightglue}~\tiny{ICCV'23} & \underline{47.9} & \underline{62.7} & \underline{73.9}  \\
    RDD+LG~\cite{lindenberger2023lightglue}~\tiny{ICCV'23} &  \textbf{55.1} & \textbf{68.9} & \textbf{78.9}  \\
    \midrule
    \textbf{\emph{Sparse with MNN}} & & &  \\
    RDD & 41.0 & 56.5 & 68.5 \\
     \bottomrule
     \end{tabular}
     \vspace{-1 em}
     \label{tab:air-ground}
 \end{table}

 \begin{figure*}
    \centering
    \includegraphics[width=0.99\linewidth]{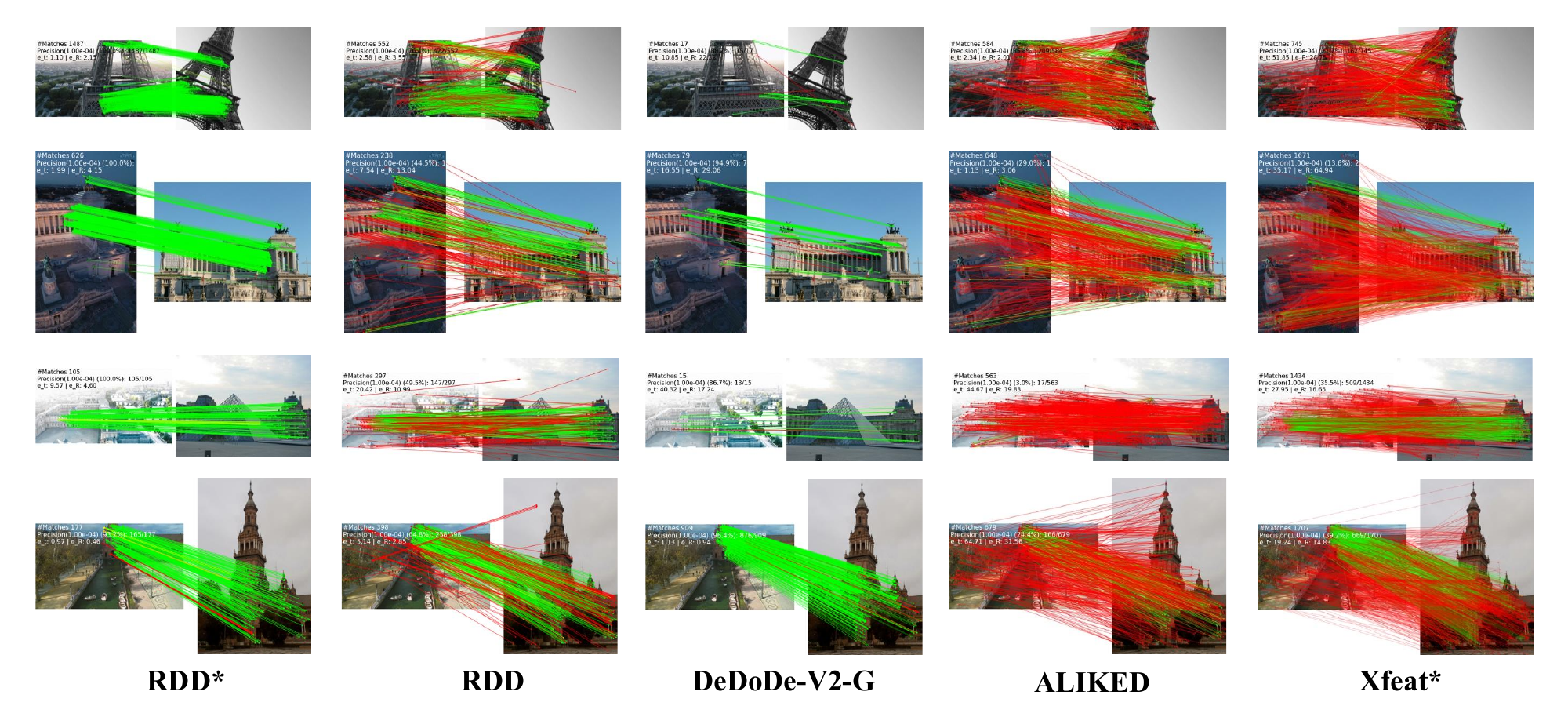}
    \caption{\textbf{Qualitative Results on Air-to-Ground.} RDD and RDD* demonstrate the ability to extract robust descriptors that perform well in cross-view settings, highlighting the effectiveness of our proposed method. DeDoDe-G* also achieves competitive performance, benefiting from the powerful foundation model Dino-v2~\cite{oquab2024dinov2}.}
    \label{fig:air}
\end{figure*}

The raw depth maps obtained from COLMAP often include significant outliers that negatively affect the accuracy of warping used to estimate overlaps and compute matching pixels. These outliers arise primarily from unmatchable moving objects and regions such as the sky or uniform foreground areas like roads. To address these challenges, we implement a series of depth post-processing steps. First, we use a semantic segmentation model~\cite{zhou2017scene} to to mask predefined classes prone to unreliable depth estimation, such as sky, sidewalks, vehicles, people, and animals and etc., which are prone to producing unreliable depth information. Second, small and isolated regions are removed using connected component analysis, discarding regions smaller than 1,000 pixels to retain only significant structures. These steps effectively enhance the quality of the depth data by mitigating noise and focusing on stable, meaningful features. 

Similar to~\cref{sec:mega-view}, for all possible aerial and ground image pairs, we apply the same warping function and threshold, and randomly select 1,500 image pairs to construct the benchmark dataset. Example pairs are shown in~\cref{fig:examples}. This benchmark dataset provides a novel air-to-ground setting for evaluating the performance of feature-matching methods.

\subsubsection{Results}
Recalling the experiment setting in Sec. 4.1 of the main paper, we report the AUC of the recovered pose under threshold (5\degree, 10\degree\ ,and 20\degree). We use RANSAC to estimate the essential matrix. We compared \shortname against the other detector/descriptor methods~\cite{superpoint2018,tyszkiewicz2020disk,Zhao2022ALIKE,zhao2023aliked,gleize2023silksimplelearned,edstedt2024dedode, potje2024cvpr}. The results are presented in~\cref{tab:air-ground} and visually in~\cref{fig:air}. Our results show a performance gain compared to previous methods. These results further confirm the robustness of RDD and RDD* in challenging scenarios.

\section{More Qualitative Results}
\cref{fig:mega1500-more} shows more qualitative results of our proposed method, RDD and RDD*, compared with other methods on MegaDepth-1500 as mentioned in the main paper, and \cref{fig:megaview-more} shows more results on MegaDepth-View. For more challenging cases, such as strong viewpoint shifts and scale changes, RDD and RDD* exhibit exceptional robustness against previous methods. This robustness is expected as our network is designed to model both geometric transformations and global context. 

\section{Running Time Analysis}
\label{sec:timing}

In this section, we present a detailed timing analysis of \shortname in both sparse and semi-dense matching settings. We also compare \shortname against other methods~\cite{superpoint2018,tyszkiewicz2020disk,Zhao2022ALIKE,zhao2023aliked,gleize2023silksimplelearned,edstedt2024dedode,potje2024cvpr}, using the same experimental settings as the main paper. All methods are evaluated on an NVIDIA RTX 4090 GPU with 24GB of VRAM. \cref{tab:time_ex} shows the inference speeds of all methods, measured in milliseconds. AUC@5\degree on MegaDepth-1500 for all methods is provided for reference. RDD and RDD* demonstrate competitive feature matching performance with competitive efficiency. A detailed breakdown of the time required for each step of our method is presented in~\cref{tab:time}. Notably, RDD is significantly faster than RDD*, as it uses fewer keypoints and does not require refinement. Although RDD* takes more time compared to \shortname, it still achieves a good balance between efficiency and performance.

\begin{table}[ht!]
    \centering
    \caption{\textbf{Runtime comparison on Megadepth-1500.} Average runtime per pair of RDD and RDD* is compared to previous methods.}
    \resizebox{0.99\columnwidth}{!}{%
    \begin{tabular}{lcc}
    \toprule
    \multirow{2}{*}{Method} & \multirow{2}{*}{Runtime (ms) $\downarrow$} & MegaDepth-1500 \\
                            &                               & (AUC @ 5\degree) $\uparrow$ \\
    \midrule
    SuperPoint~\cite{superpoint2018}~\tiny{CVPRW'18}         & 302  & 24.1 \\
    DISK~\cite{tyszkiewicz2020disk}~\tiny{NeurIPS'20}       & \underline{98}  & 38.5 \\
    ALIKED~\cite{zhao2023aliked}~\tiny{TIM'23}              & 182  & 41.8 \\
    XFeat~\cite{potje2024cvpr}~\tiny{CVPR'24}              & \textbf{32}  & 24.0 \\
    DeDoDe-G~\cite{edstedt2024dedode}~\tiny{3DV'24}        & 382 & \underline{47.2} \\
    \shortname                                               & 198 & \underline{48.2} \\
    RDD*                                                    & 416 & \textbf{51.3} \\
    \bottomrule
    \end{tabular}
    }
    \label{tab:time_ex}
\end{table}

\begin{table}[ht!]
    \centering
    \caption{\textbf{Timing Analysis.} Average required time by each step of our method on a NVIDIA RTX 4090 GPU}
    \begin{tabular}{lccc}
    \toprule
      Method & Det/Des & Matching & Refinement\\
      \midrule
       RDD  &  70 ms & 0.04 ms & - \\
       RDD* &  82 ms & 100 ms & 20 ms \\
    \bottomrule
    \end{tabular}
    \label{tab:time}
\end{table} 

\section{More Experiments}
\label{sec:experiments}
\paragraph{Better RANSAC solver} To fully understand the potential of RDD, we performed an additional experiment with better RANSAC solver Lo-RANSAC
Please see~\cref{tab:lo-ransac} for results using a better RANSAC solver. We test RDD with the same setting as the main paper.

\vspace{-1em}

\paragraph{Different Combination of Detector and Descriptor}  ~\cref{tab:combine} shows that using keypoint locations from RDD could slightly improve the performance of DeDoDe. Also, using descriptors from RDD could improve the performance of previous methods.

\begin{table}
\small
 \centering
 \caption{\textbf{RDD with LO-RANSAC}. RDD is evaluated with top 4,096 features}\vspace{-0.1em}
 \resizebox{0.99\columnwidth}{!}{%
 \begin{tabular}{lcccccc}
    \toprule      
    \multirow{3}{*}{\textbf{Method}} &\multicolumn{3}{c}{\textbf{MegaDepth-1500}} & \multicolumn{3}{c}{\textbf{MegaDepth-View}}  \\ 
     & \multicolumn{3}{c}{AUC}    &   \multicolumn{3}{c}{AUC}    \\
     & @5\degree & @10\degree & @20\degree    &      @5\degree & @10\degree & @20\degree      \\
    \midrule
    RDD & 62.9 & 75.8 & 85.1 & 61.0 & 73.4 & 81.9 \\
 \bottomrule
 \end{tabular}
 }
 \vspace{-1em}
 \label{tab:lo-ransac}
\end{table}

\begin{table}
\small
 \centering
 \caption{\textbf{Different Combination of Detector and Descriptor}. All methods are evaluated with top 4,096 features}\vspace{-0.1em}
 \resizebox{0.99\columnwidth}{!}{%
 \begin{tabular}{lcccccc}
    \toprule      
    \multirow{3}{*}{\textbf{Method}} &\multicolumn{3}{c}{\textbf{MegaDepth-1500}} & \multicolumn{3}{c}{\textbf{MegaDepth-View}}  \\ 
     & \multicolumn{3}{c}{AUC}    &   \multicolumn{3}{c}{AUC}    \\
     & @5\degree & @10\degree & @20\degree    &      @5\degree & @10\degree & @20\degree      \\
    \midrule
     RDD+DeDoDe-G~\cite{edstedt2024dedode}~\tiny{3DV'24} & 48.6 & 65.3 & 78.0 & 44.1 & 58.8 & 70.3 \\
     SuperPoint~\cite{superpoint2018}+RDD~\tiny{CVPRW'18} & 30.3 & 48.7 & 66.5 & 33.3 & 51.3 & 66.9 \\
     \midrule
     SuperPoint~\cite{superpoint2018}~\tiny{CVPRW'18} & 24.1 & 40.0 & 54.7 & 7.50 & 13.3 & 21.1 \\
    DeDoDe-V2-G~\cite{edstedt2024dedodev2,edstedt2024dedode}~\tiny{CVPRW'24, 3DV'24} & 47.2 & 63.9 & 77.5 & 33.1 & 47.6 & 60.2\\
 \bottomrule
 \end{tabular}
 }
 \vspace{-1em}
 \label{tab:combine}
\end{table}

{%
    \small%
    \bibliographystyle{ieeenat_fullname}%
    \bibliography{main}

@String(CVPR= {IEEE Conf. Comput. Vis. Pattern Recog.})

@String(ICCV= {Int. Conf. Comput. Vis.})

@String(ECCV= {Eur. Conf. Comput. Vis.})

@String(CVPRW= {IEEE Conf. Comput. Vis. Pattern Recog. Worksh.})

@String(CVPR  = {CVPR})

@String(ICCV  = {ICCV})

@String(ECCV  = {ECCV})

@String(CVPRW= {CVPRW})

@article{Lowe:2004:DIF:993451.996342,
  acmid = {996342},
  address = {Hingham, MA, USA},
  author = {Lowe, David G.},
  description = {Distinctive Image Features from Scale-Invariant Keypoints},
  doi = {10.1023/B:VISI.0000029664.99615.94},
  issn = {0920-5691},
  issue_date = {November 2004},
  journal = {Int. J. Comput. Vision},
  month = nov,
  number = 2,
  numpages = {20},
  pages = {91--110},
  publisher = {Kluwer Academic Publishers},
  title = {Distinctive Image Features from Scale-Invariant Keypoints},
  url = {http://dx.doi.org/10.1023/B:VISI.0000029664.99615.94},
  volume = 60,
  year = 2004
}

@inproceedings{sarlin20superglue,
  author    = {Paul-Edouard Sarlin and
               Daniel DeTone and
               Tomasz Malisiewicz and
               Andrew Rabinovich},
  title     = {{SuperGlue}: Learning Feature Matching with Graph Neural Networks},
  booktitle = {CVPR},
  year      = {2020},
  url       = {https://arxiv.org/abs/1911.11763}
}

@article{sun2021loftr,
  title={{LoFTR}: Detector-Free Local Feature Matching with Transformers},
  author={Sun, Jiaming and Shen, Zehong and Wang, Yuang and Bao, Hujun and Zhou, Xiaowei},
  journal={CVPR},
  year={2021}
}

@article{Huang2023adamatcher,
  title={Adaptive Assignment for Geometry Aware Local Feature Matching},
  author={Dihe Huang and Ying Chen and Yong Liu and Jianlin Liu and Shang Xu and Wenlong Wu and Yikang Ding and Fan Tang and Chengjie Wang},
  journal={{CVPR}},
  year={2023}
}

@inproceedings{wang2022matchformer,
  title={MatchFormer: Interleaving Attention in Transformers for Feature Matching},
  author={Wang, Qing and Zhang, Jiaming and Yang, Kailun and Peng, Kunyu and Stiefelhagen, Rainer},
  booktitle={Asian Conference on Computer Vision},
  year={2022}
}

@article{chen2022aspanformer,
  author    = {Chen, Hongkai and Luo, Zixin and Zhou, Lei and Tian, Yurun and Zhen, Mingmin and Fang, Tian and McKinnon, David and Tsin, Yanghai and Quan, Long},
  title     = {ASpanFormer: Detector-Free Image Matching with Adaptive Span Transformer},
  journal   = {ECCV},
  year      = {2022},
}

@inproceedings{vaswani2017attention,
  author = {Vaswani, Ashish and Shazeer, Noam and Parmar, Niki and Uszkoreit, Jakob and Jones, Llion and Gomez, Aidan N and Kaiser, {\L}ukasz and Polosukhin, Illia},
  booktitle = {Advances in Neural Information Processing Systems},
  description = {Aktuelleres Paper zur Verwendung von Attention für die Neural Machine Translation},
  pages = {5998--6008},
  title = {Attention is all you need},
  year = 2017
}

@inproceedings{li2018megadepth,
  title={{MegaDepth}: Learning single-view depth prediction from internet photos},
  author={Li, Zhengqi and Snavely, Noah},
  booktitle={CVPR},
  year={2018}
}

@inproceedings{schoenberger2016sfm,
    author={Sch\"{o}nberger, Johannes Lutz and Frahm, Jan-Michael},
    title={Structure-from-Motion Revisited},
    booktitle={Conference on Computer Vision and Pattern Recognition (CVPR)},
    year={2016},
}

@inproceedings{schoenberger2016mvs,
    author={Sch\"{o}nberger, Johannes Lutz and Zheng, Enliang and Pollefeys, Marc and Frahm, Jan-Michael},
    title={Pixelwise View Selection for Unstructured Multi-View Stereo},
    booktitle={European Conference on Computer Vision (ECCV)},
    year={2016},
}

@article{siftflow,
  author={Ce Liu and Jenny Yuen and Antonio Torralba.},
  title={{SIFT Flow}: Dense correspondence across scenes and its applications},
  journal   = {T-PAMI},
  year      = {2010},
}

@article{superpoint2018,
    author = {Daniel DeTone and Tomasz Malisiewicz and Andrew Rabinovich.},
    title = "Superpoint: Self-supervised interest point detection and description.",
    journal = "CVPR Workshops",
    year = {2018},
    pages={224-236},
}

@article{ORB2011,
    author={Ethan Rublee and Vincent Rabaud and Kurt Konolige and Gary Bradski.},
    title={ORB: An efficient alternative to SIFT or SURF},
    journal={ICCV},
    year={2011},
}

@article{SURF2008,
    author={Herbert Bay and Andreas Ess and Tinne Tuytelaars and Luc Van Gool.},
    title={Speeded-up robust features (surf)},
    journal={CVIU},
    year={2008},
    volume={110},
    number={3},
    pages={346-359},
}

@article{LIFT2016,
    author={Kwang Moo Yi and Eduard Trulls and Vincent Lepetit and Pascal Fua.},
    title={LIFT: Learned invariant feature transform},
    journal={ECCV},
    year={2016},
}

@InProceedings{edstedt2023dkm,
title={{DKM}: Dense Kernelized Feature Matching for Geometry Estimation},
author={Edstedt, Johan and Athanasiadis, Ioannis and Wadenbäck, Mårten and Felsberg, Michael},
booktitle={IEEE Conference on Computer Vision and Pattern Recognition},
year={2023}
}

@InProceedings{hpatches_2017_cvpr,
author={Vassileios Balntas and Karel Lenc and Andrea Vedaldi and Krystian Mikolajczyk},
title = {HPatches: A benchmark and evaluation of handcrafted and learned local descriptors},
booktitle = {CVPR},
year = {2017}}

@inproceedings{sarlin2019coarse,
  title     = {From Coarse to Fine: Robust Hierarchical Localization at Large Scale},
  author    = {Paul-Edouard Sarlin and
               Cesar Cadena and
               Roland Siegwart and
               Marcin Dymczyk},
  booktitle = {CVPR},
  year      = {2019}
}

@misc{oquab2024dinov2,
      title={DINOv2: Learning Robust Visual Features without Supervision}, 
      author={Maxime Oquab and Timothée Darcet and Théo Moutakanni and Huy Vo and Marc Szafraniec and Vasil Khalidov and Pierre Fernandez and Daniel Haziza and Francisco Massa and Alaaeldin El-Nouby and Mahmoud Assran and Nicolas Ballas and Wojciech Galuba and Russell Howes and Po-Yao Huang and Shang-Wen Li and Ishan Misra and Michael Rabbat and Vasu Sharma and Gabriel Synnaeve and Hu Xu and Hervé Jegou and Julien Mairal and Patrick Labatut and Armand Joulin and Piotr Bojanowski},
      year={2024},
      eprint={2304.07193},
      archivePrefix={arXiv},
      primaryClass={cs.CV}
}

@Article{kerbl3Dgaussians,
      author       = {Kerbl, Bernhard and Kopanas, Georgios and Leimk{\"u}hler, Thomas and Drettakis, George},
      title        = {3D Gaussian Splatting for Real-Time Radiance Field Rendering},
      journal      = {ACM Transactions on Graphics},
      number       = {4},
      volume       = {42},
      month        = {July},
      year         = {2023},
      url          = {https://repo-sam.inria.fr/fungraph/3d-gaussian-splatting/}
}

@article{zhao2023aliked,
  title={Aliked: A lighter keypoint and descriptor extraction network via deformable transformation},
  author={Zhao, Xiaoming and Wu, Xingming and Chen, Weihai and Chen, Peter CY and Xu, Qingsong and Li, Zhengguo},
  journal={IEEE Transactions on Instrumentation and Measurement},
  volume={72},
  pages={1--16},
  year={2023},
  publisher={IEEE}
}

@inproceedings{lindenberger2023lightglue,
  author    = {Philipp Lindenberger and
               Paul-Edouard Sarlin and
               Marc Pollefeys},
  title     = {{LightGlue: Local Feature Matching at Light Speed}},
  booktitle = {ICCV},
  year      = {2023}
}

@article{tyszkiewicz2020disk,
  title={Disk: Learning local features with policy gradient},
  author={Tyszkiewicz, Micha{\l} and Fua, Pascal and Trulls, Eduard},
  journal={Advances in neural information processing systems},
  volume={33},
  pages={14254--14265},
  year={2020}
}

@article{dusmanu2019d2net,
  title={D2-net: A trainable cnn for joint detection and description of local features},
  author={Dusmanu, Mihai and Rocco, Ignacio and Pajdla, Tomas and Pollefeys, Marc and Sivic, Josef and Torii, Akihiko and Sattler, Torsten},
  journal={arXiv preprint arXiv:1905.03561},
  year={2019}
}

@article{edstedt2024roma,
    title={{RoMa: Robust Dense Feature Matching}},
    author={Edstedt, Johan and Sun, Qiyu and Bökman, Georg and Wadenbäck, Mårten and Felsberg, Michael},
    journal={IEEE Conference on Computer Vision and Pattern Recognition},
    year={2024}
}

@inproceedings{wang2024eloftr,
  title={{Efficient LoFTR}: Semi-Dense Local Feature Matching with Sparse-Like Speed},
  author={Wang, Yifan and He, Xingyi and Peng, Sida and Tan, Dongli and Zhou, Xiaowei},
  booktitle={CVPR},
  year={2024}
}

@article{Zhao2022ALIKE,
    title = {ALIKE: Accurate and Lightweight Keypoint Detection and Descriptor Extraction},
    url = {http://arxiv.org/abs/2112.02906},
    doi = {10.1109/TMM.2022.3155927},
    journal = {IEEE Transactions on Multimedia},
    author = {Zhao, Xiaoming and Wu, Xingming and Miao, Jinyu and Chen, Weihai and Chen, Peter C. Y. and Li, Zhengguo},
    month = march,
    year = {2022},
}

@inproceedings{r2d2,
  author    = {Jerome Revaud and Philippe Weinzaepfel and C{\'{e}}sar Roberto de Souza and
               Martin Humenberger},
  title     = {{R2D2:} Repeatable and Reliable Detector and Descriptor},
  booktitle = {NeurIPS},
  year      = {2019},
}

@article{zhu2020deformable,
  title={Deformable DETR: Deformable Transformers for End-to-End Object Detection},
  author={Zhu, Xizhou and Su, Weijie and Lu, Lewei and Li, Bin and Wang, Xiaogang and Dai, Jifeng},
  journal={arXiv preprint arXiv:2010.04159},
  year={2020}
}

@misc{sun2018integralhumanposeregression,
      title={Integral Human Pose Regression}, 
      author={Xiao Sun and Bin Xiao and Fangyin Wei and Shuang Liang and Yichen Wei},
      year={2018},
      eprint={1711.08229},
      archivePrefix={arXiv},
      primaryClass={cs.CV},
      url={https://arxiv.org/abs/1711.08229}, 
}

@inproceedings{gu2021removing,
  title={Removing the Bias of Integral Pose Regression},
  author={Gu, Kang and Yang, Lei and Yao, Anbang},
  booktitle={Proceedings of the IEEE/CVF International Conference on Computer Vision},
  pages={11067--11076},
  year={2021}
}

@inproceedings{he2015deepresiduallearningimage,
  title={Deep residual learning for image recognition},
  author={He, Kaiming and Zhang, Xiangyu and Ren, Shaoqing and Sun, Jian},
  booktitle={Proceedings of the IEEE conference on computer vision and pattern recognition},
  pages={770--778},
  year={2016}
}

@inproceedings{edstedt2024dedode,
  title={{DeDoDe: Detect, Don't Describe --- Describe, Don't Detect for Local Feature Matching}},
  author = {Johan Edstedt and Georg Bökman and Mårten Wadenbäck and Michael Felsberg},
  booktitle={2024 International Conference on 3D Vision (3DV)},
  year={2024},
  organization={IEEE}
}

@INPROCEEDINGS{potje2024cvpr,
  author={Guilherme {Potje} and Felipe {Cadar} and Andre {Araujo} and Renato {Martins} and Erickson R. {Nascimento}},
  booktitle={2024 IEEE / CVF Computer Vision and Pattern Recognition (CVPR)}, 
  title={XFeat: Accelerated Features for Lightweight Image Matching}, 
  year={2024}}

@article{luo2020aslfeat,
  title={ASLFeat: Learning Local Features of Accurate Shape and Localization},
  author={Luo, Zixin and Zhou, Lei and Bai, Xuyang and Chen, Hongkai and Zhang, Jiahui and Yao, Yao and Li, Shiwei and Fang, Tian and Quan, Long},
  journal={Computer Vision and Pattern Recognition (CVPR)},
  year={2020}
}

@inproceedings{gleize2023silksimplelearned,
  title={Silk: Simple learned keypoints},
  author={Gleize, Pierre and Wang, Weiyao and Feiszli, Matt},
  booktitle={Proceedings of the IEEE/CVF international conference on computer vision},
  pages={22499--22508},
  year={2023}
}

@InProceedings{li2022decoupling,
    title={Decoupling Makes Weakly Supervised Local Feature Better},
    author={Li, Kunhong and Wang, Longguang and Liu, Li and Ran, Qing and Xu, Kai and Guo, Yulan},
    booktitle = {Proceedings of the IEEE/CVF Conference on Computer Vision and Pattern Recognition (CVPR)},
    month     = {June},
    year      = {2022},
    pages     = {15838-15848}
}

@misc{rocco2018neighbourhoodconsensusnetworks,
      title={Neighbourhood Consensus Networks}, 
      author={Ignacio Rocco and Mircea Cimpoi and Relja Arandjelović and Akihiko Torii and Tomas Pajdla and Josef Sivic},
      year={2018},
      eprint={1810.10510},
      archivePrefix={arXiv},
      primaryClass={cs.CV},
      url={https://arxiv.org/abs/1810.10510}, 
}

@inproceedings{chen2024geometryawarefeaturematchinglargescale,
  title={Geometry-aware feature matching for large-scale structure from motion},
  author={Chen, Gonglin and Wu, Jinsen and Chen, Haiwei and Teng, Wenbin and Gao, Zhiyuan and Feng, Andrew and Qin, Rongjun and Zhao, Yajie},
  booktitle={2025 International Conference on 3D Vision (3DV)},
  pages={34--43},
  year={2025},
  organization={IEEE}
}

@misc{lin2018focallossdenseobject,
      title={Focal Loss for Dense Object Detection}, 
      author={Tsung-Yi Lin and Priya Goyal and Ross Girshick and Kaiming He and Piotr Dollár},
      year={2018},
      eprint={1708.02002},
      archivePrefix={arXiv},
      primaryClass={cs.CV},
      url={https://arxiv.org/abs/1708.02002}, 
}

@misc{loshchilov2019decoupledweightdecayregularization,
      title={Decoupled Weight Decay Regularization}, 
      author={Ilya Loshchilov and Frank Hutter},
      year={2019},
      eprint={1711.05101},
      archivePrefix={arXiv},
      primaryClass={cs.LG},
      url={https://arxiv.org/abs/1711.05101}, 
}

@inproceedings{hu2019local,
  title={Local relation networks for image recognition},
  author={Hu, Han and Zhang, Zheng and Xie, Zhenda and Lin, Stephen},
  booktitle={Proceedings of the IEEE/CVF international conference on computer vision},
  pages={3464--3473},
  year={2019}
}

@misc{sattler2018benchmarking6dofoutdoorvisual,
      title={Benchmarking 6DOF Outdoor Visual Localization in Changing Conditions}, 
      author={Torsten Sattler and Will Maddern and Carl Toft and Akihiko Torii and Lars Hammarstrand and Erik Stenborg and Daniel Safari and Masatoshi Okutomi and Marc Pollefeys and Josef Sivic and Fredrik Kahl and Tomas Pajdla},
      year={2018},
      eprint={1707.09092},
      archivePrefix={arXiv},
      primaryClass={cs.CV},
      url={https://arxiv.org/abs/1707.09092}, 
}

@misc{liu2019giftlearningtransformationinvariantdense,
      title={GIFT: Learning Transformation-Invariant Dense Visual Descriptors via Group CNNs}, 
      author={Yuan Liu and Zehong Shen and Zhixuan Lin and Sida Peng and Hujun Bao and Xiaowei Zhou},
      year={2019},
      eprint={1911.05932},
      archivePrefix={arXiv},
      primaryClass={cs.CV},
      url={https://arxiv.org/abs/1911.05932}, 
}

@misc{dosovitskiy2021imageworth16x16words,
      title={An Image is Worth 16x16 Words: Transformers for Image Recognition at Scale}, 
      author={Alexey Dosovitskiy and Lucas Beyer and Alexander Kolesnikov and Dirk Weissenborn and Xiaohua Zhai and Thomas Unterthiner and Mostafa Dehghani and Matthias Minderer and Georg Heigold and Sylvain Gelly and Jakob Uszkoreit and Neil Houlsby},
      year={2021},
      eprint={2010.11929},
      archivePrefix={arXiv},
      primaryClass={cs.CV},
      url={https://arxiv.org/abs/2010.11929}, 
}

@inproceedings{mishkin2018repeatabilityenoughlearningaffine,
  title={Repeatability is not enough: Learning affine regions via discriminability},
  author={Mishkin, Dmytro and Radenovic, Filip and Matas, Jiri},
  booktitle={Proceedings of the European conference on computer vision (ECCV)},
  pages={284--300},
  year={2018}
}

@article{ono2018lfnetlearninglocalfeatures,
  title={LF-Net: Learning local features from images},
  author={Ono, Yuki and Trulls, Eduard and Fua, Pascal and Yi, Kwang Moo},
  journal={Advances in neural information processing systems},
  volume={31},
  year={2018}
}

@article{jaderberg2016spatialtransformernetworks,
  title={Spatial transformer networks},
  author={Jaderberg, Max and Simonyan, Karen and Zisserman, Andrew and others},
  journal={Advances in neural information processing systems},
  volume={28},
  year={2015}
}

@inproceedings{barrosolaguna2019keynetkeypointdetectionhandcrafted,
  title={Key. net: Keypoint detection by handcrafted and learned cnn filters},
  author={Barroso-Laguna, Axel and Riba, Edgar and Ponsa, Daniel and Mikolajczyk, Krystian},
  booktitle={Proceedings of the IEEE/CVF international conference on computer vision},
  pages={5836--5844},
  year={2019}
}

@article{mishchuk2018workinghardknowneighbors,
  title={Working hard to know your neighbor's margins: Local descriptor learning loss},
  author={Mishchuk, Anastasiia and Mishkin, Dmytro and Radenovic, Filip and Matas, Jiri},
  journal={Advances in neural information processing systems},
  volume={30},
  year={2017}
}

@inproceedings{katharopoulos2020transformersrnnsfastautoregressive,
  title={Transformers are rnns: Fast autoregressive transformers with linear attention},
  author={Katharopoulos, Angelos and Vyas, Apoorv and Pappas, Nikolaos and Fleuret, Fran{\c{c}}ois},
  booktitle={International conference on machine learning},
  pages={5156--5165},
  year={2020},
  organization={PMLR}
}

@inproceedings{zhou2017scene,
    title={Scene Parsing through ADE20K Dataset},
    author={Zhou, Bolei and Zhao, Hang and Puig, Xavier and Fidler, Sanja and Barriuso, Adela and Torralba, Antonio},
    booktitle={Proceedings of the IEEE Conference on Computer Vision and Pattern Recognition},
    year={2017}
}

@inproceedings{edstedt2024dedodev2,
  title={{DeDoDe v2: Analyzing and Improving the DeDoDe Keypoint Detector
}},
  author = {Johan Edstedt and Georg Bökman and Zhenjun Zhao},
  booktitle={IEEE/CVF Computer Society Conference on Computer Vision and Pattern Recognition Workshops (CVPRW)},
  year={2024},
}
}

\end{document}